# Realistic noise synthesis reduces bias and improves tissue microstructure estimation with supervised machine learning

Bradley G. Karat[1,2*], Maëliss Jallais[3,4*], Ali R. Khan[1], Santiago Aja-Fernández[5,6], Jelle Veraart[2], Marco Palombo[3,4]

[1]Robarts Research Institute, Schulich School of Medicine and Dentistry, The University of Western Ontario, London, Canada

[2]Center for Biomedical Imaging, Department of Radiology, NYU Grossman School of Medicine, New York, New York, USA

[3]Cardiff University Brain Research Imaging Center (CUBRIC), School of Psychology, Cardiff University, Cardiff, United Kingdom

[4]School of Computer Science and Informatics, Cardiff University, Cardiff, United Kingdom

[5]Laboratorio de Procesado de Imagen, Universidad de Valladolid, Valladolid, Spain

[6]Instituto de Investigación Biosanitaria de Valladolid, IBioVALL, Valladolid, Spain

*Shared first author

## Abstract

**Purpose:** Diffusion MRI (dMRI) enables non-invasive probing of tissue microstructure, but accurate parameter estimation is challenged by noise-related effects. Magnitude reconstruction produces Rician-distributed signals whose expectation deviates from the true signal at low SNR, while preprocessing further alters the effective noise characteristics. In supervised machine learning (ML) frameworks that rely on simulated training data, discrepancies between the noise characteristics of simulated and acquired signals introduce a form of covariate shift, whereby the input signal distribution differs between training and inference. We investigated the impact of this mismatch on microstructure parameter estimation and propose a realistic noise synthesis (RNS) strategy to mitigate it.

**Methods:** We developed a two-step RNS framework that incorporates both the Rician expectation and the effective post-processing noise variance into simulated training signals. The Rician expectation was modelled using a noise standard deviation estimated with Marchenko–Pastur Principal Component Analysis (MPPCA), while the effective standard deviation was derived from spherical harmonic residuals of preprocessed data. The method was evaluated using a two-compartment cylinder–zeppelin model and the SANDI model on simulated datasets across multiple SNR levels and on in vivo diffusion data with repeated acquisitions. Sensitivity to noise misestimation was additionally assessed.
**Results:** Training on signals that ignored magnitude-induced effects produced systematic, SNR-dependent parameter bias, particularly at low SNR. Incorporating the Rician expectation substantially reduced bias to the level of noise-aware nonlinear least-squares fitting. Additionally modelling the effective standard deviation further improved precision. Performance was largely independent of regression architecture but sensitive to accurate noise estimation.
**Conclusion:** Realistic modelling of noise statistics in simulated training data mitigates signal-domain covariate shift and is essential for unbiased supervised microstructure estimation, particularly in low-SNR regimes associated with high b-values or high spatial resolution.

## 1. Introduction

The non-invasive mapping of tissue microstructure using MRI holds the promise to transform our understanding of both natural physiological and diseased states. Diffusion MRI (dMRI) has emerged as a popular technique to estimate tissue properties non-invasively as it is sensitive to the ubiquitous diffusion of water, which given the measurement is on a micrometer scale[1–4]. Indeed, dMRI has seen utility in a variety of applications including: stroke[5–7], brain development[8–10], tumor assessment[11–14], multiple sclerosis[15–17], epilepsy[18–20], and neurodegeneration[21–24]. While promising, it is a challenging inverse problem to infer relevant tissue properties from the dMRI signal[25–28]. One source of systematic bias in dMRI tissue parameter estimation arises from the non-trivial noise properties of magnitude MR data[29–31].

Current state-of-the-art approaches address this problem either with noise-aware model fitting, such as non-linear least squares (NLLS) or maximum-likelihood estimation, which can reduce bias but are often computationally expensive and sensitive to model complexity, or with machine learning (ML) estimators, which are fast and precise but often neglect realistic noise modelling during training, limiting generalisation and introducing bias[32–34].

This study aims to determine how magnitude-induced bias and post-processing noise variance affect both conventional and supervised ML estimation methods, and to develop a realistic noise synthesis strategy that better matches training data to experimental signals in order to improve the accuracy and robustness of microstructure parameter estimation in dMRI.

Noise in MR measurements primarily refers to thermal noise arising from the receiver coils. This noise is well modeled as zero-mean, white Gaussian noise in the complex-valued signal measured by each receive coil. In the absence of additional reconstruction processing, this complex noise is commonly assumed to be stationary and is characterized by a single parameter, the noise standard deviation $\sigma$ [31,35–37]. However, modern MRI acquisitions typically employ multichannel receiver coils together with parallel imaging reconstruction methods, such as SENSE[38] or GRAPPA[39], which modify the variance of the noise differently for each spatial position. As a result, the noise variance becomes non-stationary, i.e., spatially dependent. Additional processing steps performed in the scanner may further introduce spatial correlations and departures from ideal theoretical noise models.

MR data are most often reconstructed and analyzed in magnitude form, a nonlinear operation that alters the complex Gaussian noise statistics of the signal. In practice, despite the additional complexities introduced by parallel imaging and scanner-specific reconstruction procedures, the single-coil magnitude signal can be accurately approximated by a Rician distribution (noncentral chi distribution for multicoil magnitude reconstructions), parameterized by a spatially varying $\sigma$ value. Importantly, the expectation of the magnitude signal does not coincide with the true underlying signal amplitude $S$, particularly at low signal-to-noise ratio (SNR) (approximately <5), where the expected magnitude signal exceeds $S$ [29,30,40].

While this description applies to the reconstructed magnitude signal, common image preprocessing steps, and particularly denoising methods such as Marchenko–Pastur Principal Component Analysis (MPPCA)[37], further modify the empirical signal distribution. These methods reconstruct each voxel as a weighted combination of neighboring measurements, effectively performing a local estimation of the expected signal value through spatial and/or measurement averaging[41,42]. Consequently, by the central limit theorem, the post-processed signal distribution is often well approximated by a non-zero-mean Gaussian. However, because most generic denoising methods implicitly assume Gaussian-distributed data, they effectively estimate the expectation of the magnitude distribution rather than the true underlying signal amplitude $S$. Thus, although preprocessing substantially reduces the variance of the noise, the estimated signal remains shifted relative to $S$, yielding a Gaussian-approximated signal with reduced standard deviation $\sigma^*$ but mean $\mu_1(S, \sigma)$ inherited from the underlying magnitude distribution, as illustrated in Figure 1.

This effect has been shown to lead to systematic errors in MR parameter estimation, as model fitting is performed on signals whose expectation does not correspond to the true underlying signal amplitude. These effects are particularly pronounced in diffusion MRI, where sensitivity to microscopic water motion and tissue microstructure is achieved through diffusion weighting, which inherently reduces signal amplitude and increases the prevalence of low-SNR measurements[32,43,44].

Although SNR can be improved through hardware and acquisition strategies, such as higher magnetic field strengths, optimized receiver coils, or signal averaging, these approaches increase cost and scan time and may exacerbate motion-related artifacts. As a more practical alternative, post-processing denoising methods have been extensively developed and applied to diffusion MRI data[45], explicitly accounting for noise-related bias by incorporating estimates of the noise standard deviation (referred to here as $\hat{\sigma}$) into the parameter estimation framework. For example, Jones and Basser[32] embedded a noise parameter into NLLS estimation of the apparent diffusion coefficient (ADC), yielding an estimator that approximates the expectation of a Rician-distributed signal, and Wiest-Daesslé et al.[46] adapted the non-local means filter[47] to Rician noise corrupted data. Maximum-likelihood approaches have also been employed for estimating parameters from biophysical models and signal representations by maximizing the likelihood of the Rician-distributed dMRI signal given the model[48–54]. While statistically principled, such approaches are often computationally demanding and sensitive to model complexity and initialization.

Recently, ML-based approaches have gained popularity for estimating MR tissue parameters. Compared with conventional optimization-based methods, ML estimators can offer substantially faster inference, improved robustness in sparsely sampled acquisition schemes, and low computational cost once trained. ML methods have been applied to estimate a wide range of parameters, including myelin water fraction[55], T1 and T2 relaxation times[56,57], magnetization transfer metrics[58], and diffusion microstructure parameters derived from multi-compartment biophysical models[33,59–63]. These approaches are commonly categorized as either self-supervised or supervised. Self-supervised methods estimate parameters by minimizing discrepancies between measured signals and model-predicted signals, without requiring explicit ground-truth parameter labels. In contrast, supervised methods learn a direct mapping from measured signals to parameters using labeled training data, typically generated via simulations.

While supervised approaches can exhibit systematic bias if the training data are not representative of experimental conditions, they have been shown to achieve higher precision and lower variance than self-supervised methods in many practical settings, which is often advantageous for studies focused on group differences, effect sizes, and statistical power[34]. At the same time, prior work has demonstrated that the performance of supervised ML estimators depends critically on the realism of the training distribution[64,65] and training strategy[66,67]. In particular, realistic modeling of noise, accounting for both the magnitude-induced bias and the effective noise standard deviation after preprocessing, is frequently neglected. As a result, training data may fail to reflect experimental conditions, leading to reduced generalization and biased parameter estimates in practice.

In this study, we investigate how noise characteristics, and specifically magnitude-induced Rician bias and post-processing standard deviation, affect microstructure parameter estimation in both conventional NLLS fitting and supervised ML frameworks, with effects that are particularly pronounced in low-SNR regimes. We introduce a two-step realistic noise synthesis (RNS) strategy that incorporates both the Rician expectation and the effective noise standard deviation into simulated training data, thereby aligning the statistical properties of the training and inference signals. Using both simulated and in vivo datasets, we demonstrate that explicitly modeling these noise characteristics substantially reduces SNR-dependent bias in supervised ML estimators, bringing their accuracy to the level of noise-aware NLLS approaches while further improving estimation precision. These results highlight the importance of realistic noise modeling for robust and unbiased microstructure inference in diffusion MRI.

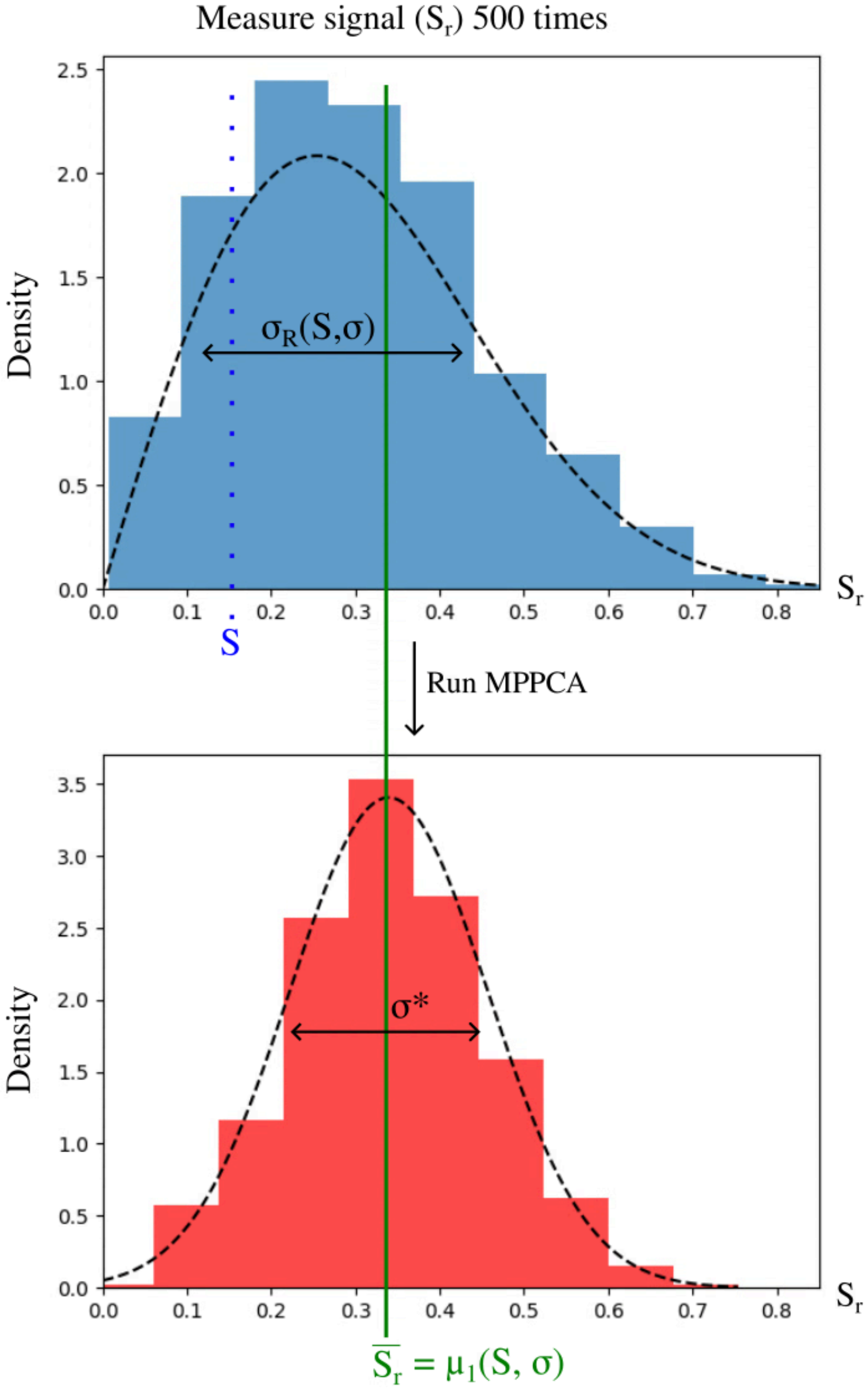

**Figure 1.** Simulation illustrating how preprocessing alters the empirical signal distribution while preserving magnitude-induced bias. A true noiseless signal amplitude (S) corrupted by thermal noise with standard deviation $\sigma$ was used to simulate 500 noisy magnitude measurements at low SNR ($\sim N(5, 0.5)$) within a large patch of 729 voxels (see *Section 2.4*). For a representative voxel at the center of the patch and an arbitrary diffusion-encoding direction, the distribution of the raw measurements follows a Rician distribution with mean $\mu_1(S,\sigma)$ and standard deviation $\sigma_R(S,\sigma) = \sqrt{2\sigma^2 + S^2 - \mu_1(S,\sigma)}$. MPPCA [37], a commonly used denoising method in dMRI preprocessing pipelines, was then applied to the patch. After denoising, the signal distribution at the same voxel and diffusion direction is well approximated by a Gaussian distribution with a mean of $\mu_1(S,\sigma)$, but a reduced standard deviation $\sigma^* < \sigma$, demonstrating variance reduction without removal of magnitude-related bias.

## 2. Methods

In this section we present our proposed two-step RNS framework that models both the Rician mean offset and the post-processing standard deviation to make supervised dMRI training data statistically closer to the signals encountered at inference. We present application to multi-compartment biophysical signal models for microstructure imaging with dMRI, focusing on two representative spherical-mean signal models of increasing complexity. We evaluated our approach on simulated datasets across multiple SNR levels and on in vivo repeated-acquisition data, comparing against noise-aware NLLS and alternative supervised regressors trained with the noise models currently used for supervised ML inference in this field.

### *2.1. Multi-compartment biophysical signal models*

Consider an arbitrary biophysical model of diffusion consisting of $M$ compartments. The dMRI signal $S$ can be written as

$$S(b, \hat{\boldsymbol{g}}; \boldsymbol{\Theta}) = \sum_{m=1}^{M} f_m S_m(b, \hat{\boldsymbol{g}}; \boldsymbol{\theta}_m) , \quad (1)$$

where $b$ denotes the diffusion weighting (the b-value), $\hat{\boldsymbol{g}}$ the diffusion-encoding direction, $f_m$ the signal fraction of compartment $m$ with $\sum_{m=1}^{M} f_m = 1$, $\boldsymbol{\theta}_m$ the biophysical parameters describing compartment $m$ (e.g. orientation and parallel diffusivity for a stick compartment), and $S_m$ the corresponding compartment-specific signal model. This forward model maps a parameter set $\boldsymbol{\Theta} = \{f_m, \boldsymbol{\theta}_m\}_{m=1}^{M}$ to a noiseless signal $S$.

Such forward models form the basis for generating training data in supervised ML applications, where $\boldsymbol{\Theta}$ is typically sampled densely within biologically plausible bounds and the corresponding noiseless signals are computed for a fixed acquisition protocol. Alternative strategies for constructing training datasets include Monte Carlo simulations of complex synthetic substrates[54,61,68], digital phantoms[69–71], finite element methods[72] or parameter maps derived from previously acquired data fitted using NLLS methods[73,74]. In all fully supervised cases, an ML model is trained to learn a mapping from the signal $S$ to the parameters $\boldsymbol{\Theta}$.

However, because experimentally measured magnitude MR signals are affected by noise-induced bias, particularly at low SNR, noiseless simulated signals $S$ can be a poor approximation of the real measured signal $S_r$ (Figure 2B). As a result, supervised ML models trained on noiseless signals may fail to generalize to experimental data, leading to biased parameter estimates at inference[75,76]. Similar issues arise in self-supervised approaches if noise-related biases are not explicitly accounted for in the loss function[33].

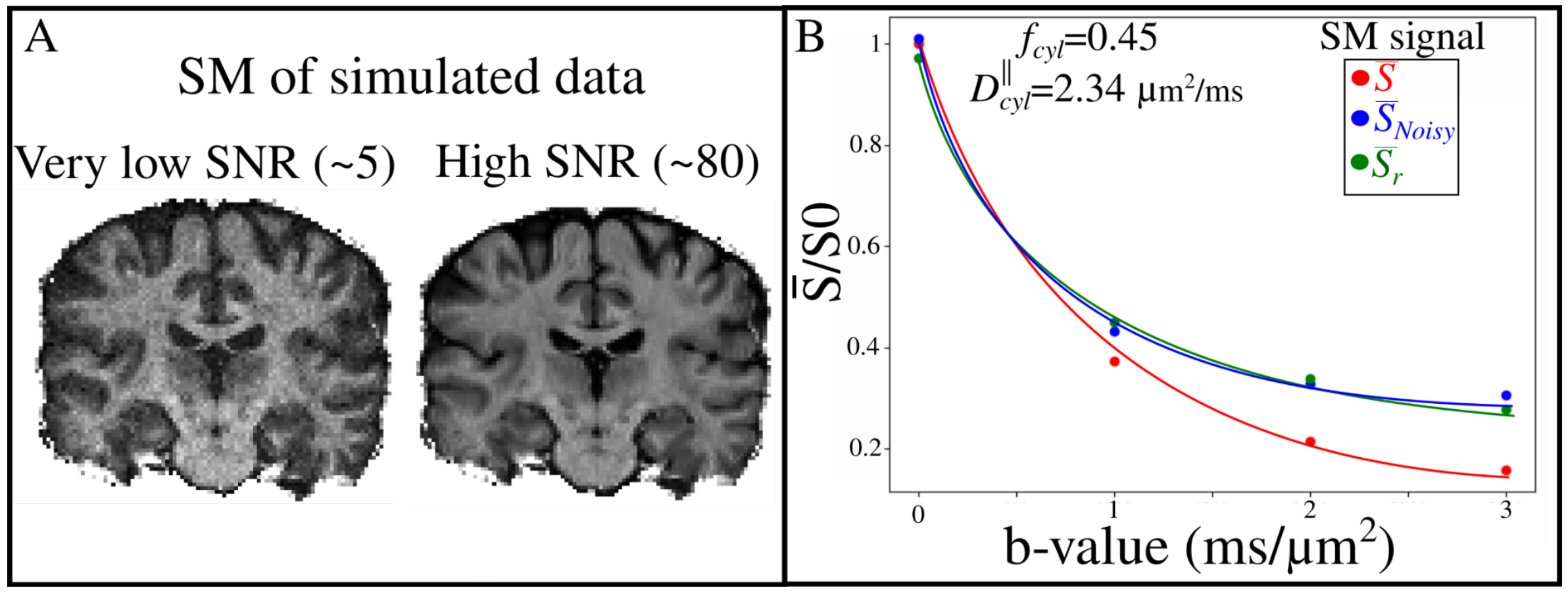

**Figure 2.** Simulated normalized spherical mean (SM) signal generated using the acquisition parameters of the 1.5 mm³ dataset (Section 2.4). (A) Coronal slice of simulated SM signals at $b = 3$ ms/μm² shown for a very low and high SNR scenarios using the proposed Realistic Noise Synthesis (RNS) method (*Section 2.4*). (B) SM signal at a representative voxel. Green points denote the measured signal $\bar{S}_r$ from the real data, with estimated parameters $f_{cyl} = 0.45$ and $D^{||}_{cyl} = 2.34$ μm²/ms. Red points show the corresponding noiseless simulated SM signal $\bar{S}$ (Eq. 11) generated using these estimated parameters. Blue points represent the noisy simulated SM signal $\bar{S}_{Noisy}$ (Eq. 7), obtained by injecting noise using the proposed RNS approach, using both the MPPCA-derived standard deviation $\hat{\sigma}$, and the spherical harmonics (SH)-derived $\hat{\sigma}*$. Solid lines indicate exponential fits.

*2.2. Proposed Realistic Noise Synthesis (RNS) method*

To address the mismatch between simulated and experimental signals, we propose a Realistic Noise Synthesis (RNS) framework that first learns the noise statistics directly from the real data and then injects noise into simulated training signals in a manner consistent with both magnitude-induced bias and the effective noise standard deviation observed after preprocessing. Importantly and differently from any current supervised learning approach, our RNS is data-conditioned, that is built on empirical noise distributions estimated directly from the target dataset instead of generated from arbitrary priors. The method consists of three key steps: (0) Estimating voxel-wise $\sigma$ and $\sigma^*$ from the masked brain data; (1) modelling the magnitude-induced bias by replacing the noiseless signal with the expectation of a Rician distribution with standard deviation $\sigma$; and (2) adding Gaussian noise with standard deviation $\sigma^*$ to capture the effective post-processing noise level.

A crucial feature of the framework is that, rather than using a single scalar noise level or random synthetic noise priors, RNS reproduces the observed distribution of noise variance present in the inference data, including their spatial heterogeneity after preprocessing. The resulting simulations are therefore statistically closer to the acquired signals in both mean and variance, which is precisely the mechanism by which RNS mitigates signal-domain covariate shift. To our knowledge, this dataset-conditioned, empirical sampling strategy is distinct from the standard practice in supervised ML for dMRI inference, where training signals are typically corrupted using simplified noise models that include the experimental statistics, but are not specifically designed to mirror them.

**Step 0: Learning empirical noise distributions**

After restricting the data to voxels within the brain mask, voxel-wise estimates of the thermal noise standard deviation $\hat{\sigma}$ and the effective post-processing noise standard deviation $\hat{\sigma}^*$ are computed and aggregated across the whole brain. These pooled values define empirical distributions from which noise parameters are randomly sampled during training-data synthesis in Steps 1 and 2.

The thermal noise standard deviation $\sigma$ is estimated before other preprocessing steps, for example using MPPCA[37], and denoted $\hat{\sigma}$.

To estimate the effective noise standard deviation $\sigma^*$ present in pre-processed experimental data, we fit spherical harmonics (SH) independently to each b-value shell of the measured signal $S_r$, truncated at a maximum harmonic order $l_{max}$:

$$S_r(b, \hat{\boldsymbol{g}}) = \sum_{l=0}^{lmax} \sum_{m=-l}^{l} c_l^m(b) Y_l^m(\hat{\boldsymbol{g}}) \tag{4}$$

where $c_l^m(b)$ are the SH coefficients, $Y_l^m(\hat{\boldsymbol{g}})$ the SH basis functions, and $l_{max}$ is determined by the number of gradient directions acquired for the given b-value[77]. Note that only even orders of $l$ are used, under the assumption of antipodally symmetric diffusion[78].

After estimating the SH coefficients, we compute the residuals between the SH reconstruction and the measured diffusion signal:

$$\varepsilon = \mathbf{CY} - S_r \tag{5}$$

where $\mathbf{C}$ and $\mathbf{Y}$ denote the matrices of SH coefficients and basis functions, respectively. Because the residuals include not only thermal noise but also potential model truncation and preprocessing artefacts, we summarize their dispersion using the median absolute deviation (MAD), and convert it to a robust estimate of the effective post-processing noise standard deviation $\sigma^*$ for each b-value shell:

$$\hat{\sigma}_b^* = 1.4826 \sqrt{\frac{N_b}{N_b - M_b}}\, MAD(\varepsilon) \tag{6}$$

Here, $N_b$ denotes the number of gradient directions in shell $b$, and $M_b$ the number of fitted SH coefficients in that shell, which depends on the chosen truncation order $l_{max}$ (with only even orders included). The factor $\sqrt{\frac{N_b}{N_b - M_b}}$ corrects for the loss of degrees of freedom introduced by the SH fit, and the factor 1.4826 converts the MAD to the standard deviation under the assumption of normally distributed residuals. The final estimate of $\sigma^*$, denoted $\hat{\sigma}^*$, is obtained by averaging $\hat{\sigma}_b^*$ across all b-values, yielding a more robust and stable estimate under the assumption that the effective post-processing noise standard deviation is independent of b-value (see Results section).

**Step 1: Injection of magnitude-induced bias**

Given a noiseless signal $S(b, \hat{\boldsymbol{g}}; \boldsymbol{\Theta})$ generated from the forward model (Eq. 1) and an estimate of the thermal noise standard deviation $\hat{\sigma}$, we compute the expectation of the corresponding Rician distribution:

$$\mu_1(S(b, \hat{\boldsymbol{g}}; \boldsymbol{\Theta}), \hat{\sigma}) = \hat{\sigma}\sqrt{\frac{\pi}{2}}\, L_{1/2}\left(-\frac{S(b, \hat{\boldsymbol{g}}; \boldsymbol{\Theta})^2}{2\hat{\sigma}^2}\right), \tag{2}$$

where $L_q(\cdot)$ denotes the Laguerre polynomial of order $q$. For computational convenience, we use the identity

$$L_{\frac{1}{2}}(x) = e^{\frac{x}{2}}\left[(1 - x) I_0\left(-\frac{x}{2}\right) - x I_1\left(-\frac{x}{2}\right)\right], \tag{3}$$

with $I_0$ and $I_1$ denoting the modified Bessel functions of the first kind of order 0 and 1, respectively. This step accounts for the magnitude-induced bias present in experimentally measured signals, particularly at low SNR.

**Step 2: Applying the effective noise standard deviation after preprocessing**

Finally, we generate the noisy simulated signal as

$$S_{Noisy}(b, \hat{\boldsymbol{g}}; \boldsymbol{\Theta}) = \mu_1(S(b, \hat{\boldsymbol{g}}; \boldsymbol{\Theta}), \hat{\sigma}) + N(0, \hat{\sigma}^*) \tag{7}$$

where $N(0, \hat{\sigma}^*)$ is a random sample from a normal distribution with zero mean and standard deviation $\hat{\sigma}^*$, independently sampled for each b-value. The resulting $S_{Noisy}$ is Gaussian distributed, with a mean governed by $\mu_1$, the expectation of the Rician distribution, and a standard deviation $\hat{\sigma}^*$, consistent with that observed in preprocessed experimental data (Figure 1). By construction, $S_{Noisy}$ more closely approximates the measured signal $S_r$ than noiseless simulations (Figure 2B), or noisy signal generated using a uniform distribution of noise variances between arbitrary ranges, including the experimental ranges. Consequently, supervised ML models trained using $S_{Noisy}$ are expected to generalize better to experimental data and yield improved parameter estimates at inference.

The code used for the current study is available at https://github.com/Bradley-Karat/RNS_dMRI_ML. As well, we provide a BIDS application which implements the proposed framework for fitting a variety of models at https://github.com/Bradley-Karat/Supervised_RNS_dMRI.

*2.3 Biophysical models used for evaluation*

To demonstrate and evaluate the proposed RNS framework, we considered two biophysical diffusion models of increasing complexity, both expressed using the spherical mean (SM) signal representation. The SM formulation removes orientation dependence by averaging the diffusion-weighted signal over gradient directions, allowing model behavior to be examined independently of fiber orientation dispersion.

**Two-compartment cylinder-zeppelin model**

As a low-dimensional example, we considered a two-compartment model consisting of a zero-radius cylinder ("stick") and a zeppelin[79]. The cylinder represents an anisotropic compartment characterized by a unit orientation $\hat{\boldsymbol{w}} = (\phi, \psi)$, a parallel diffusivity $D_{cyl}^{\parallel}$, negligible perpendicular diffusivity, and a signal fraction $f_{cyl}$. The zeppelin is modeled as an anisotropic Gaussian compartment with parallel and perpendicular diffusivities $D_{zep}^{\parallel}$ and $D_{zep}^{\perp}$, respectively.

The diffusion signal $S(b, \hat{\boldsymbol{g}}; \boldsymbol{\Theta})$ of this two-compartment model is defined as:

$$S(b, \hat{\boldsymbol{g}}; \boldsymbol{\Theta}) = f_{cyl}\, S_{cyl}\big(b, \hat{\boldsymbol{g}}; \hat{\boldsymbol{w}}, D_{cyl}^{\parallel}\big) + \big(1 - f_{cyl}\big)\, S_{zep}\big(b, \hat{\boldsymbol{g}}; \hat{\boldsymbol{w}}, D_{zep}^{\parallel}, D_{zep}^{\perp}\big), \tag{8}$$

where $\boldsymbol{\Theta} = \{f_{cyl}, \hat{\boldsymbol{w}}, D_{cyl}^{\parallel}, D_{zep}^{\parallel}, D_{zep}^{\perp}\}$ and the compartment signals are given by:

$$S_{cyl}\big(b, \hat{\boldsymbol{g}}; \hat{\boldsymbol{w}}, D_{cyl}^{\parallel}\big) = e^{-b\,\langle \hat{\boldsymbol{w}}, \hat{\boldsymbol{g}} \rangle^2 D_{cyl}^{\parallel}}, \tag{9}$$

$$S_{zep}\big(b, \hat{\boldsymbol{g}}; \hat{\boldsymbol{w}}, D_{zep}^{\parallel}, D_{zep}^{\perp}\big) = e^{-b\,\left(\langle \hat{\boldsymbol{w}}, \hat{\boldsymbol{g}} \rangle^2 D_{zep}^{\parallel} + \left(1 - \langle \hat{\boldsymbol{w}}, \hat{\boldsymbol{g}} \rangle^2\right) D_{zep}^{\perp}\right)}. \tag{10}$$

The corresponding SM signal can be written as:

$$\frac{\bar{S}(b; \boldsymbol{\Theta})}{S_0} = f_{cyl} \frac{\sqrt{\pi}\, erf\left(\sqrt{bD_{cyl}^{||}}\right)}{2\sqrt{bD_{cyl}^{||}}} + (1 - f_{cyl}) e^{-bD_{zep}^{\perp}} \frac{\sqrt{\pi}\, erf\left(\sqrt{b(D_{zep}^{||} - D_{zep}^{\perp})}\right)}{2\sqrt{b(D_{zep}^{||} - D_{zep}^{\perp})}} \tag{11}$$

where $\bar{S}(b)$ is the direction-averaged diffusion signal, $S_0$ is the signal without diffusion weighting, and $erf(\cdot)$ is the error function. In this implementation, to reduce the dimensionality of the model, we assume that the parallel diffusivities of the two compartments are equal ($D_{\mathrm{zep}}^{\parallel} = D_{\mathrm{cyl}}^{\parallel}$), and impose a tortuosity constraint on the zeppelin perpendicular diffusivity $D_{\mathrm{zep}}^{\perp} = \big(1 - f_{\mathrm{cyl}}\big) D_{\mathrm{cyl}}^{\parallel}$ [80], as done by Gyori et al.[64]. Under these constraints, only two independent parameters are estimated: the cylinder signal fraction $f_{\mathrm{cyl}}$, and the parallel diffusivity $D_{\mathrm{cyl}}^{\parallel}$.

**SANDI model**

To assess the proposed method under a higher-order biophysical model, we additionally considered the Soma and Neurite Density Imaging (SANDI) model[60]. SANDI requires high b-values to achieve sensitivity to restricted diffusion within soma, resulting in substantial signal attenuation and reduced SNR at the highest diffusion weightings. These acquisition conditions place parameter estimation in a low-SNR regime, where noise-induced bias is expected to have a greater impact, thereby providing a suitable test case for evaluating the proposed noise synthesis framework.

The SM signal representation of SANDI is given by:

$$\frac{\bar{S}(b; \boldsymbol{\Theta})}{S_0} = f_{\mathrm{neurite}}\, \bar{S}_{\mathrm{n}}(b; D_{\mathrm{n}}) + f_{\mathrm{soma}}\, \bar{S}_{\mathrm{s}}(b; r_{\mathrm{s}}, D_{\mathrm{n}}) + (1 - f_{\mathrm{neurite}} - f_{\mathrm{soma}})\, \bar{S}_{\mathrm{ec}}(b; D_{\mathrm{ec}}) \tag{12}$$

where $\bar{S}_{\mathrm{n}}$, $\bar{S}_{\mathrm{s}}$, and $\bar{S}_{\mathrm{ec}}$ are the SM signal contributions from the intraneurite, intrasoma, and extracellular compartments, respectively, $f_{\mathrm{neurite}}$ the intraneurite signal fraction, $f_{\mathrm{soma}}$ the soma signal fraction, $D_\mathrm{n}$ and $D_\mathrm{ec}$ the intraneurite and extracellular diffusivities, and $r_s$ the soma radius, leading to $\mathbf{\Theta} = \{f_\mathrm{neurite}, f_\mathrm{soma}, D_\mathrm{n}, r_\mathrm{s}, D_\mathrm{ec}\}$.

Model fitting was performed using a Python implementation adapted from the original MATLAB SANDI toolbox (available at https://github.com/Bradley-Karat/Supervised_RNS_dMRI). Parameter maps analysed here include $f_{neurite}$, $f_{soma}$ and the soma radius $r_s$.

*2.4 dMRI data*

We evaluated the proposed method using dMRI datasets from Manzano Patrón et al.[45] and from a separate ultra–high-gradient acquisition[81], selected to match the requirements of the biophysical models considered in this study and to probe performance across different SNR regimes.

For the two-compartment SM model described in *Section 2.3*, we used the multi-resolution datasets from Manzano Patrón et al.[45], which provide repeated measurements and moderate b-values suitable for estimating axonal signal fraction and parallel diffusivity. Two spatial resolutions are available (1.5 mm³ and 0.9 mm³), providing datasets with different intrinsic SNR levels. The 1.5 mm³ dataset exhibits relatively high SNR (~59), limiting its suitability for evaluating noise-induced bias in parameter estimation. We therefore used this dataset to obtain whole-brain distributions of $f_\mathrm{cyl}$ and $D_\mathrm{cyl}^{\parallel}$, which were subsequently treated as reference values for generating simulated diffusion data at varying SNR levels, enabling a quantitative evaluation of the proposed noise synthesis framework (*Section 2.5*). In contrast, the 0.9 mm³ dataset has lower intrinsic SNR due to its higher spatial resolution and was used to qualitatively assess the performance of the proposed method under more challenging noise conditions.

The 1.5 mm³ isotropic dataset (TR = 3.23 s, TE = 89.2 ms), consists of five repeats with 300 volumes each. Of these, 297 volumes were acquired with anterior–posterior (AP) phase encoding, including 27 non–diffusion-weighted volumes ($b = 0$ ms/µm²) and 90 volumes at $b = 1, 2,$ and 3 ms/µm², while three additional $b = 0$ ms/µm² volumes were acquired with posterior–anterior (PA) phase encoding. The 0.9 mm³ isotropic dataset (TR = 3.0 s, TE = 92 ms) consists of four repeats with 116 volumes each. The repeated acquisitions enabled comparison of parameter estimates from a single acquisition versus the average across four repeats, thereby providing an empirical evaluation of SNR effects in vivo. For both datasets, preprocessed magnitude data were used (see Manzano Patron et al.[45] for preprocessing steps).

To evaluate the proposed framework using the SANDI model, we used diffusion MRI data from a single subject acquired on a 3T Siemens Connectom system equipped with ultra-strong gradients (maximum gradient strength 300 mT/m)[81]. The diffusion data were acquired at 2 $mm^3$ isotropic resolution (TR = 3.0 s, TE = 59 ms) with b-values of 0 (14 volumes), 0.5 (30 directions), 1.2 (30 directions), 2.4 (60 directions), 4.0 (60 directions), and 6.0 (60 directions) $ms/\mu m^2$. Preprocessing details for this dataset are described in Genc et al.[81].

*2.5 Simulated dMRI data*

To quantitatively assess the error and bias of the proposed method under controlled noise conditions, we simulated dMRI data at multiple SNR levels, closely mimicking real data. To obtain a spatially varying noise distribution, we generated a voxel-wise noise standard deviation $\sigma$ on a grid matching the 1.5 $mm^3$ dataset described above (*Section 2.4*). For each voxel, $\sigma$ was sampled from a normal distribution with mean of 1/SNR and variance of 10% of the mean (1/(SNR*10)), with SNR levels of 10 (very low), 20 (low), 40 (medium), and 80 (high). A Gaussian smoothing kernel with a full width at half maximum of 4 mm was applied to get a smooth but spatially varying noise map within the brain at each SNR. This spatial map is referred to as the ground-truth $\sigma$. The distribution of this ground-truth $\sigma$ is shown in Supplementary Figure S1.

To obtain realistic tissue parameters for the simulations, we fitted the SM two-compartment cylinder-zeppelin model to the acquired data (see *Section 2.5*) with a standard NLLS estimator implemented in the DMIPY toolbox (version 1.0.1)[82], and estimated $f_{cyl}$ and $D_{cyl}^{||}$. These voxel-wise estimates were subsequently used as ground truth values for our simulations.

The simulation process for each voxel $i$ was as follows:

1. Sample a random orientation $\hat{\boldsymbol{w}}$ for the cylinder by drawing spherical angles $\phi \in [0, \pi]$ and $\psi \in [-\pi, \pi]$.
2. Simulate the noiseless directional signal $S(b, \hat{\boldsymbol{g}}; \boldsymbol{\Theta})$ (Eq. 8) using the acquisition scheme of the 1.5 $mm^3$ real data (*Section 2.4*), the sample orientation $\hat{\boldsymbol{w}}$, and the reference values $f_{cyl}$ and $D_{cyl}^{||}$ in that voxel.

3. For each SNR level, add complex Gaussian noise to the noiseless signal. Specifically, for each b-value, compute the magnitude signal

$$S_{Noisy}(b, \hat{\boldsymbol{g}}; \boldsymbol{\Theta}) = \sqrt{\mathrm{Re}(S(b, \hat{\boldsymbol{g}}; \boldsymbol{\Theta}))^2 + \mathrm{Im}(S(b, \hat{\boldsymbol{g}}; \boldsymbol{\Theta}))^2},$$

   with:

$$\mathrm{Re}(S(b, \hat{\boldsymbol{g}}; \boldsymbol{\Theta})) = S(b, \hat{\boldsymbol{g}}; \boldsymbol{\Theta}) + \varepsilon_r,$$

$$\mathrm{Im}(S(b, \hat{\boldsymbol{g}}; \boldsymbol{\Theta})) = \varepsilon_i,$$

   where $\varepsilon_r \sim N(0, \sigma)$ and $\varepsilon_i \sim N(0, \sigma)$ are independent Gaussian noise terms.
4. For each SNR level, compute the spherical mean signal at each b-value by averaging the magnitude signal across all gradient directions and normalize it with respect to the $b = 0$ ms/μm$^2$ signal.

To get a pragmatic estimate of the noise $\hat{\sigma}$, we performed MPPCA[37] on the simulated directional data (obtained at step 3) using only the $b = 1$ ms/μm$^2$. We then applied MPPCA denoising to the full data, and used this denoised signal for the experiments. The resulting normalized spherical mean signal across SNR and b-value is shown in Supplementary Figure S2.

*2.6 Diffusion parameter estimators*

We investigated the fitting results using NLLS and supervised learning approaches, considering multiple strategies for incorporating noise into the training signal set. Evaluation was performed on a simulated test dataset (Section 2.5), to quantify the bias and error in the estimated parameters across SNR levels, and real data (Section 2.4), for a comparison of parameter maps across estimators.

The first considered approach, referred to as *Estimator 1*, is a conditional least square estimator[43] that accounts for the Rician bias of magnitude-reconstructed diffusion data. At each voxel, the parameter vector $\boldsymbol{\Theta}$ is estimated as:

$$\widehat{\boldsymbol{\Theta}} = \arg\min_{\boldsymbol{\Theta}} \frac{1}{N} \sum_{b=1}^{N} [\bar{S}(b; \boldsymbol{\Theta}, \mu_1(S(\mathrm{b}, \hat{\boldsymbol{g}}; \boldsymbol{\Theta}), \hat{\sigma})) - \bar{S}_r(b)]^2 , \quad (13)$$

where $N$ is the number of b-values, $S(b, \hat{\boldsymbol{g}}; \boldsymbol{\Theta})$ is the simulated directional signal given the forward model, $\hat{\sigma}$ is the estimated noise standard deviation in that voxel, $\mu_1$ is the Rician mean of the simulated directional signal given $\hat{\sigma}$, $\bar{S}$ is the simulated SM signal, and $\bar{S}_r$ is the SM of the measured signal. This estimator was implemented in DMIPY[82].

For the supervised approach, we used a bootstrap-aggregating regressor (BR) with 200 decision trees, using a mean squared error loss as the fit criterion, implemented using scikit-learn (version 1.5.0)[83]. We simulated 100,000 noise-free SM signals (Eq. 8; similarly to steps 1 and 2 in *Section 2.5*) by uniformly sampling the parameters on biologically realistic intervals, i.e. $f_{cyl} \in [0.01, 0.99]$ and $D_{cyl}^{||} \in [0.01, 3]$ ms/μm$^2$. Different noise injection strategies were then used to generate the training datasets. In the first variant, referred to as *Estimator 2a*, Rician offset computed using $\hat{\sigma}$ and added to the noiseless signals (*Section 2.2* - Step 1).
In the second variant, referred to as *Estimator 2b*, zero-mean Gaussian noise with standard deviation $\hat{\sigma}^*$ estimated from the corresponding inference dataset (*Section 2.2* - Step 3), was added to the noiseless signals. In *Estimator 2c*, the proposed RNS method was applied, including both the Rician expectation with standard deviation $\hat{\sigma}$ and the Gaussian noise with standard deviation $\hat{\sigma}^*$ (*Section 2.2*). For each simulated signal, $\hat{\sigma}$ and $\hat{\sigma}^*$ were randomly sampled (with replacement) from within the brain mask of the dataset used for evaluation. This produces training signals that reflect both the Rician-induced offset and the standard deviation characteristics observed in the measured data. To isolate the impact of noise misestimation, a fourth variant, *Estimator 2d*, used the same procedure as *Estimator 2c* but replaced $\hat{\sigma}$ with the ground truth $\sigma$ (available for simulated datasets only). Additional experiments were conducted to assess the sensitivity of the method to misestimation of the noise level, by introducing 10% and 20% under- and overestimation of the noise standard deviation.

To examine whether performance differences were dependent on the regression architecture, we additionally trained multi-layer perceptron (MLP) regressors implemented in scikit-learn[83], with three hidden layers of 30 neurons each, ReLU activation functions, and the Adam optimizer[84]. Three variants were considered. *Estimator 3a* was trained on simulated signals with a Rician offset computed using $\hat{\sigma}$, *Estimator 3b* on signals with zero-mean Gaussian noise with standard deviation $\hat{\sigma}^*$, and *Estimator 3c* was trained using the proposed noise synthesis framework, using $\hat{\sigma}$ and $\hat{\sigma}^*$. In the three cases, the training datasets were identical to those used for *Estimator 2a*, *2b* and *2c*, respectively.

## 3. Results

### *3.1. Validation of the SH-based noise estimate*

We first assessed whether the SH-based estimate $\hat{\sigma}^*$ accurately reflects the standard deviation of the post-processed magnitude signal. Using controlled simulations, we compared the empirically measured standard deviation of the processed signals with the corresponding SH-based estimates. As shown in Figure 3, $\hat{\sigma}^*$ closely matches the observed standard deviation across SNR levels, supporting its use as an estimate of the effective post-processing noise standard deviation.

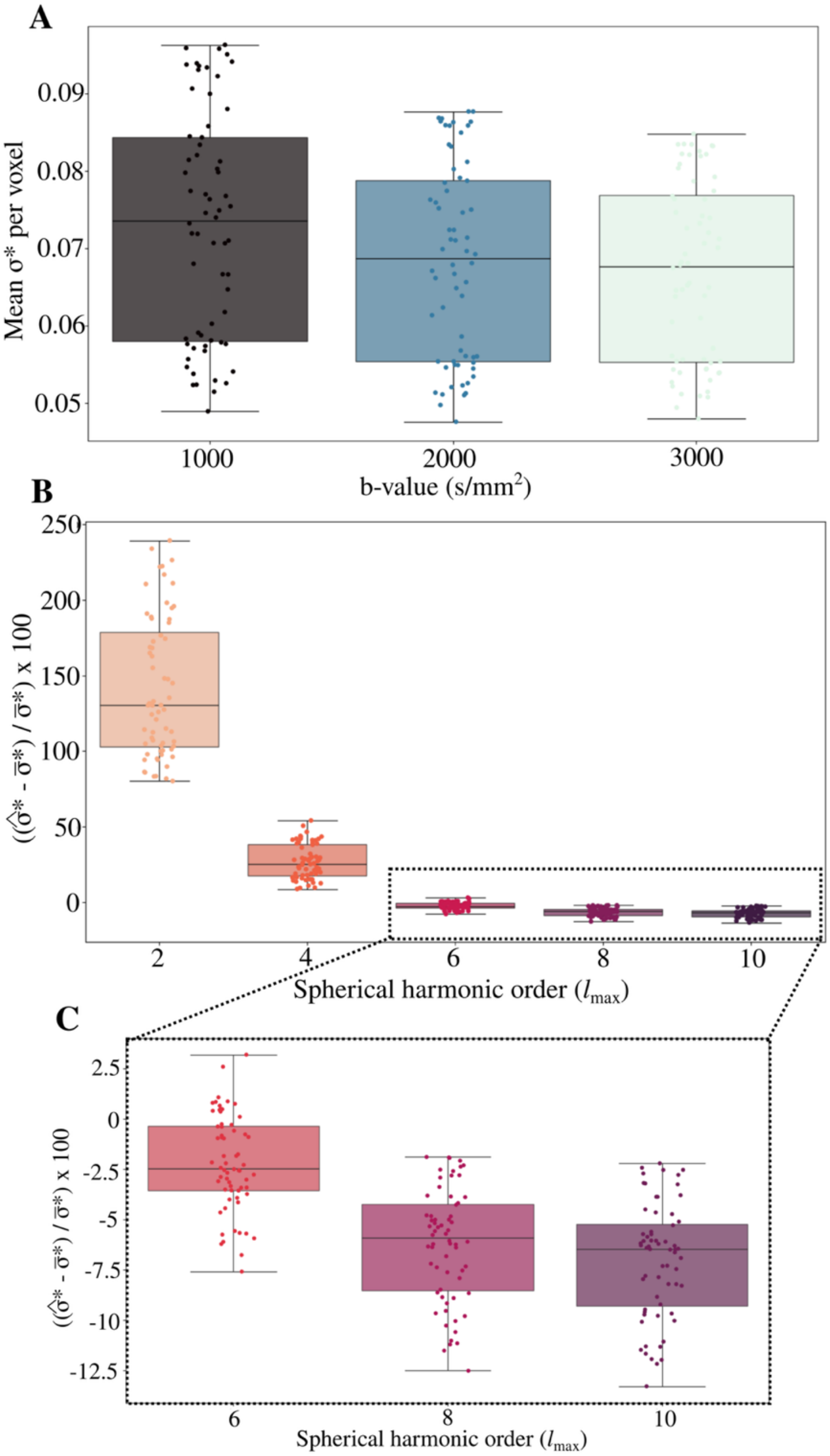

**Figure 3.** Validation of the SH-based estimate ($\hat{\sigma}^*$) of the effective post-processing noise standard deviation ($\sigma^*$). As in Figure 1, complex Gaussian noise with known standard deviation $\sigma$ was added to noiseless simulated directional signals, followed by magnitude reconstruction and MPPCA denoising. For each voxel and b-value shell, multiple independent realizations were generated. The empirical standard deviation of the post-processed magnitude signal was computed across realizations and compared to the SH-residual-based estimate $\hat{\sigma}^*$. (A) The mean empirical $\sigma^*$ across gradient directions for each b-value and plotted across a patch of 64 voxels. (B) The percent difference between the SH-based estimate ($\hat{\sigma}^*$) and the mean empirical $\sigma^*$ ($\bar{\sigma}^*$; averaged across gradient directions and b-values) for varying spherical harmonic orders ($l_{\max}$) across the 64 voxels. (C) The inset image zooms in on the harmonic orders from 6 to 10.

*3.2 Effect of SNR and noise modeling on parameter estimation*

Figure 4 shows the estimated $f_{cyl}$ maps from the simulated dataset in the very low (~10) and high SNR (~80) levels, using *Estimators 1&2* (*Section 2.5*). At high SNR, all estimators recover the ground-truth distribution with good visual correspondence. However, as SNR decreases, clear differences emerge. *Estimator 2b*, trained on signals with zero-mean Gaussian noise, exhibits reduced tissue contrasts and poor correspondence with ground truth at very low SNR. In contrast, incorporating the Rician-induced offset improves anatomical contrast, as can be seen with *Estimator 1* and *2a*, with substructures such as the thalamus becoming conspicuous. Further qualitative improvement is observed when both the Rician expectation and noise standard deviation are modeled during training (*Estimators 2c* and *2d*), with clearer white matter-grey matter delineation and better agreement with the ground truth. The use of the true noise standard deviation (*Estimator 2d*) further reduces visible discrepancies relative to *Estimator 2c*.

To quantify these observations, we evaluated the residuals of the fitted parameters within the white matter, where such a two-compartment model would find the most robust use. Figure 5 shows the distribution of the $f_{cyl}$ residuals for each estimator across SNR. The mean residual, referred to as the bias here, gives a measure of accuracy of the estimators, while the standard deviation quantifies the precision of the estimates. At high SNR, all estimators are accurate and precise. As SNR decreases, substantial bias appears when the Rician-induced offset is not accounted for (*Estimator 2b*), with bias magnitude increasing as noise increases. Incorporating the Rician expectation (*Estimator 1*, *2a*, *2c* and *2d*) largely removes systematic bias. When both the Rician offset and standard deviation are modeled during training (*Estimators 2c* and *2d*), the standard deviation of residuals is reduced, indicating improved precision. Using the ground-truth noise standard deviation $\sigma$ (*Estimator 2d*) provides a further reduction in variance relative to using the estimated $\hat{\sigma}^*$ (*Estimator 2c*). Supplementary Figure S3 shows the same results with similar trends for $D_{cyl}^{||}$.

The structure of the bias in parameter space is shown in Figure 6. At low SNR, we can observe large systematic deviations from the ground truth, as shown by the large magnitude of the arrows. The magnitude of these deviations decreases with increasing SNR. Estimators incorporating the proposed RNS (*2c* and *2d*) show consistently smaller bias vectors across parameter space. We can also note that bias is higher at extreme values of $f_{\mathrm{cyl}}$ and $D_{\mathrm{cyl}}^{\parallel}$.

Differences between *Estimators 2c* and *2d* arise from noise standard deviation misestimation, as the fitting process is identical except for the Rician mean step. Supplementary Figure S1 shows the distributions of the ground-truth $\sigma$ and estimated $\hat{\sigma}$ noise standard deviations within the simulated brain. It can be seen that across SNR, MPPCA consistently underestimates the ground-truth $\sigma$, in agreement with previous observations[37]. This underestimation propagates to the final parameter estimation, resulting in a higher residual bias, particularly visible at lower SNR.

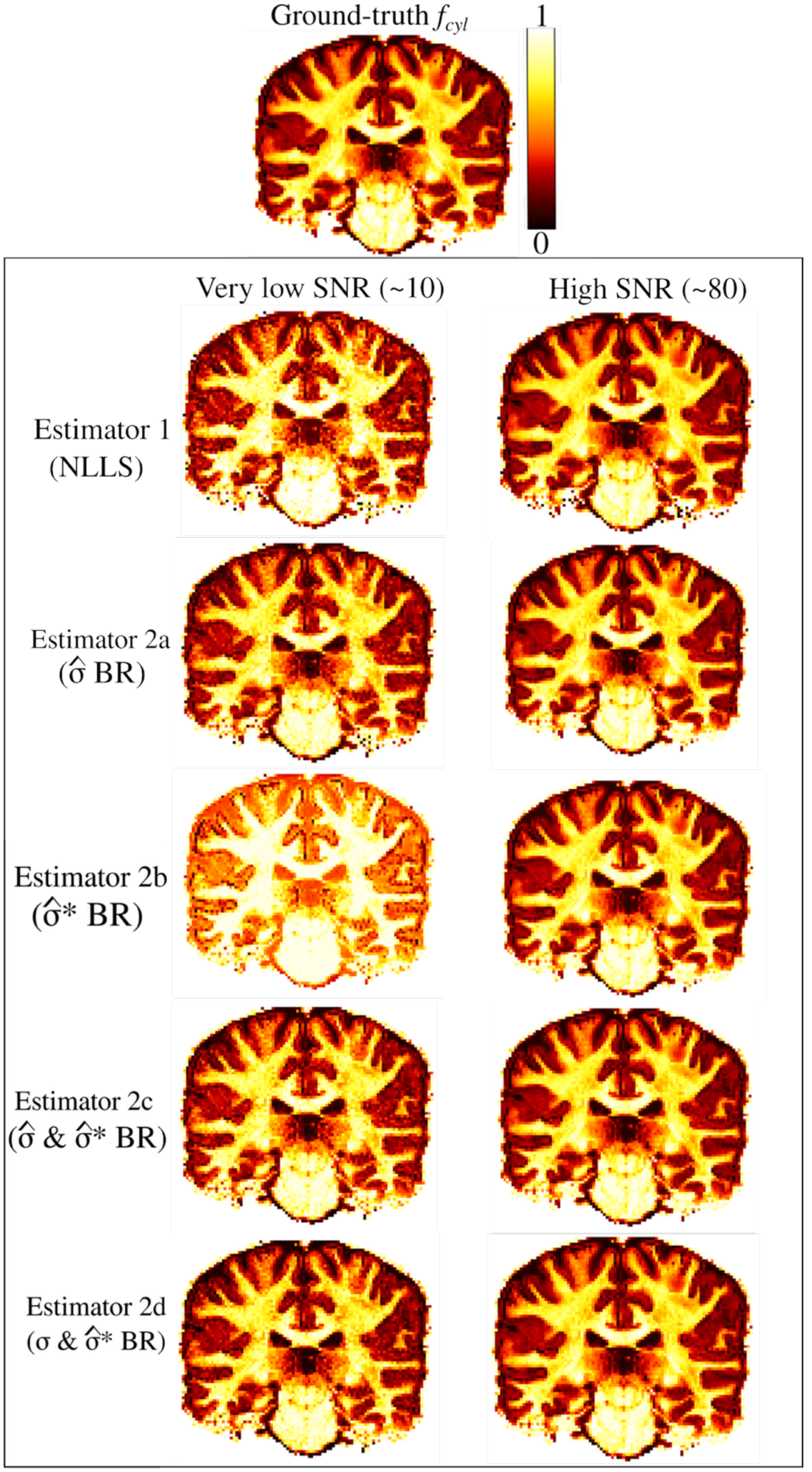

**Figure 4.** Estimated maps from the simulated dataset for two SNR levels (~10 and 80) for the different estimators. The top row shows the ground-truth map. *Estimator 1* corresponds to NLLS fitting with a Gaussian offset; *Estimators 2a* & *2b* are the baseline regressors trained on signals Rician offset and zero-mean Gaussian noise respectively; *Estimator 2c* incorporates the proposed RNS using an estimated noise standard deviation; *Estimator 2d* uses the ground-truth noise standard deviation. At high SNR, all estimators closely approximate the ground truth. At lower SNR, differences in contrast and anatomical correspondence become apparent, particularly for *Estimator 2b* that does not explicitly model Rician noise effects. Color bars are identical across panels.

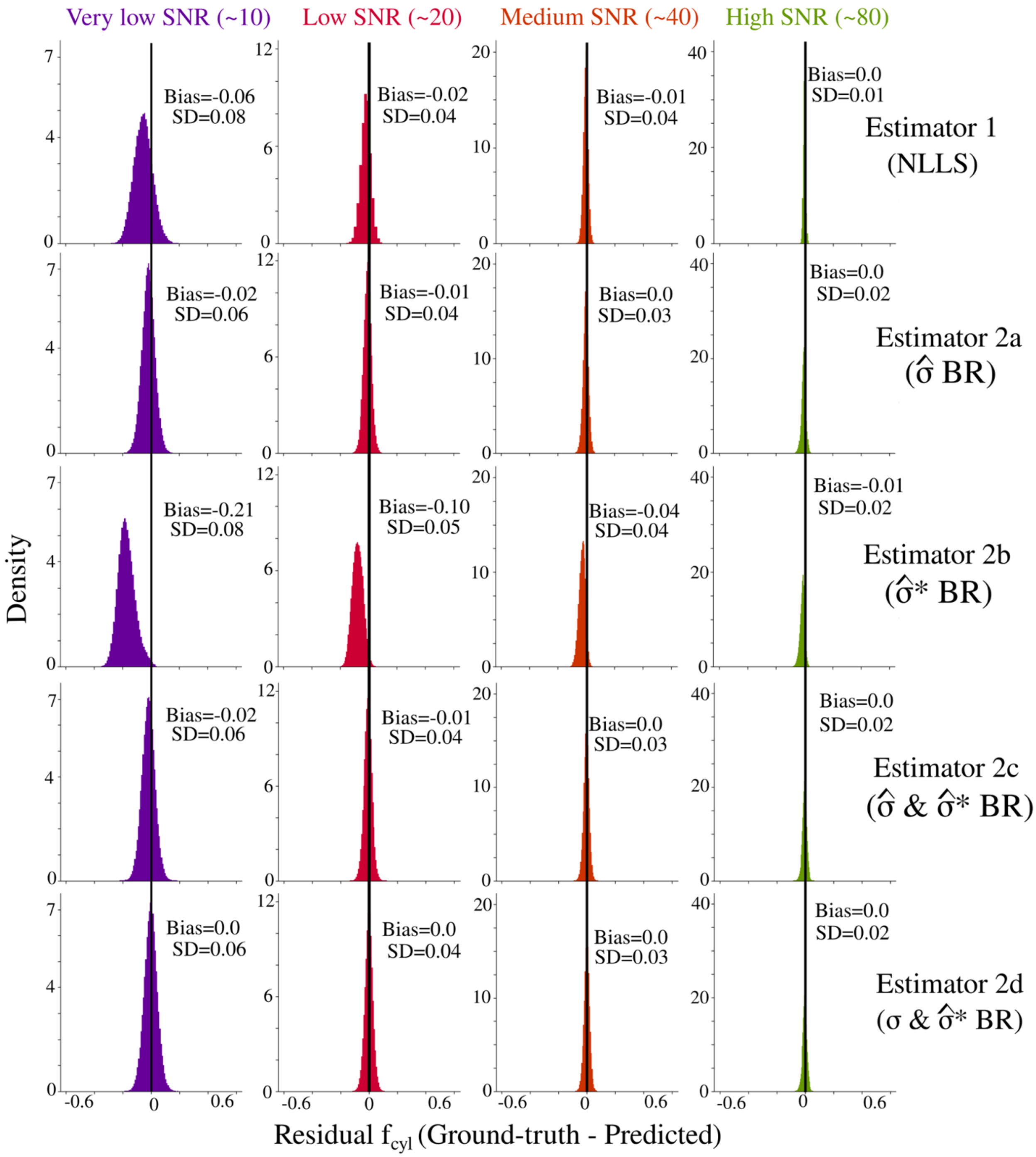

**Figure 5.** Distribution of $f_{\text{cyl}}$ residuals (ground truth minus estimate) within the white matter across SNR levels for each estimator. Columns correspond to increasing SNR (~10, 20, 40, 80), and rows to estimators. The vertical black line indicates zero residual. Reported values denote the mean residual (bias) and standard deviation (SD). At high SNR, all estimators exhibit low bias and variance. At lower SNR, substantial bias emerges when Rician effects are not modeled (*Estimator 2b*), whereas estimators incorporating Rician noise modeling (*Estimators 1*, *2a*, *2c*, *2d*) maintain near-zero bias.

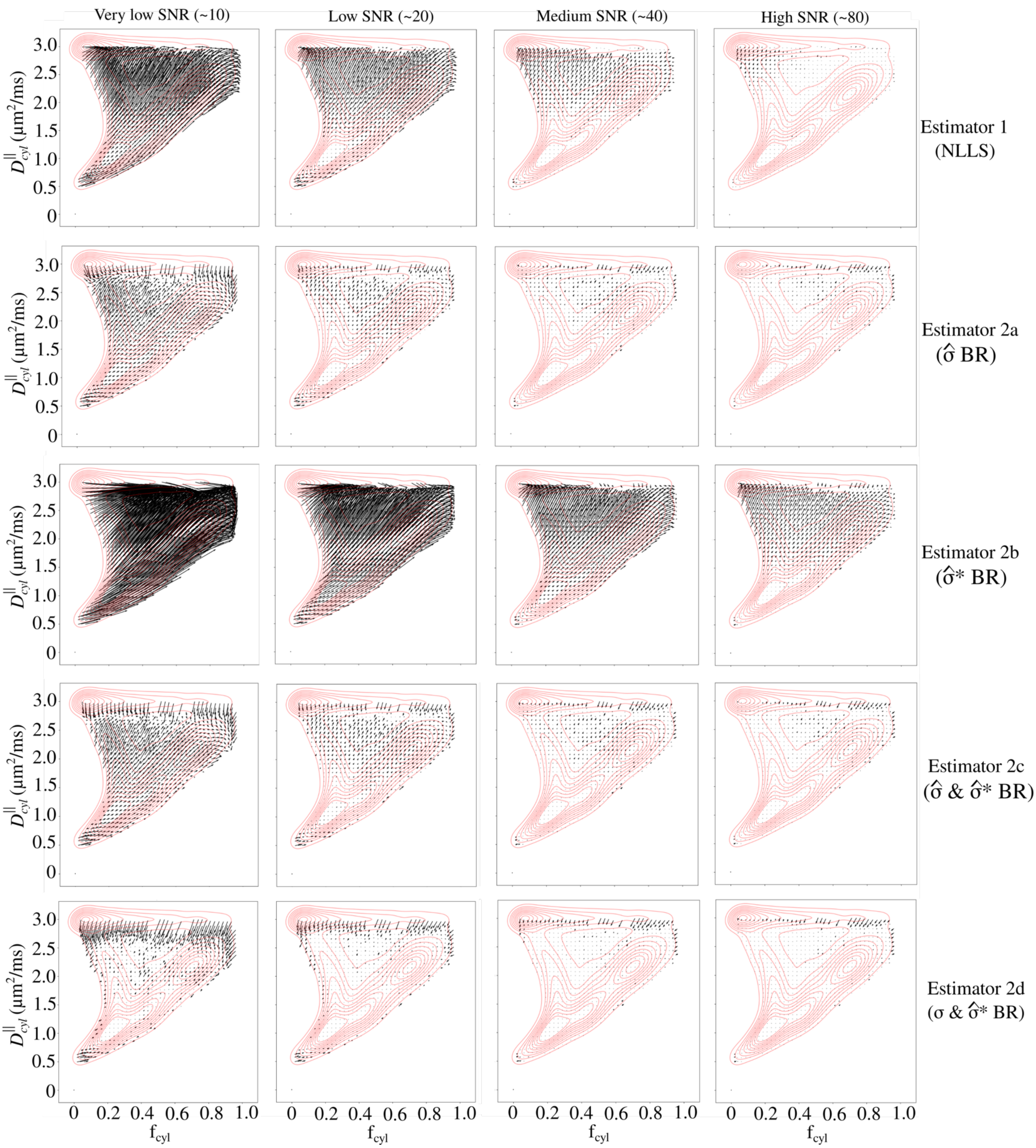

**Figure 6.** Bias in the joint estimation of $f_{\text{cyl}}$ and $D_{\text{cyl}}^{\|}$ shown as quiver plots across SNR levels (columns) and estimators (rows). Arrows represent the mean deviation from the ground-truth

parameters to the mean estimated values over the simulated parameter grid. Larger arrow magnitudes indicate greater systematic bias. Red contours indicate the density of the ground-truth parameter distribution across the whole brain. Bias increases with decreasing SNR and is most pronounced when the Rician offset is not incorporated during training (*Estimator 2b*). Estimators using the proposed Rician noise synthesis (*2c* and *2d*) show reduced bias magnitude across parameter space.

### *3.3 Impact of noise misestimation on parameter estimation*

To evaluate the sensitivity of the proposed framework to inaccuracies in noise estimation, we performed additional experiments using *Estimator 1* and *2c*. During training, the noise level used in the Rician expectation step was intentionally mis-specified. Specifically, the estimated noise standard deviation $\hat{\sigma}$ was scaled relative to the ground-truth value $\sigma$ as $\hat{\sigma} = \alpha\sigma$, with $\alpha \in \{0.8, 0.9, 1.0, 1.1, 1.2\}$, corresponding to ±10% and ±20% under- and overestimation. All other components of the pipeline were kept unchanged. For *Estimator 1*, the Rician correction applied to the data before model fitting used the same $\hat{\sigma}$ as specified above.

Figure 7 shows the distribution of residuals for $f_{\mathrm{cyl}}$ within the white matter across SNR levels under these conditions. When the noise level is correctly specified ($\hat{\sigma} = \sigma$), both estimators exhibit no bias (same results as *Estimator 2d* in Figure 5), with the dispersion of the residuals increasing as SNR decreases. In contrast, systematic bias arises when the noise standard deviation is misestimated. Underestimation of the noise ($\hat{\sigma} < \sigma$) leads to a negative residual bias, corresponding to an overestimation of $f_{\mathrm{cyl}}$, whereas overestimation ($\hat{\sigma} > \sigma$) results in a positive bias, i.e., an underestimation of $f_{\mathrm{cyl}}$. The magnitude of this bias increases with the degree of misestimation and is more pronounced at lower SNR. In comparison, the effect of noise misestimation on the dispersion of the residuals is relatively limited, with only minor changes in standard deviation observed across conditions.

A comparison between the NLLS estimator and *Estimator 2c* shows that NLLS exhibits smaller mean bias across noise misestimation levels and SNR regimes. However, it is characterized by a broader distribution of residuals, indicating reduced precision and a higher proportion of voxels with large estimation errors.

Overall, these results demonstrate that accurate estimation of the noise standard deviation is critical for unbiased parameter recovery, as errors in $\hat{\sigma}$ propagate into systematic deviations in the estimated microstructural parameters.

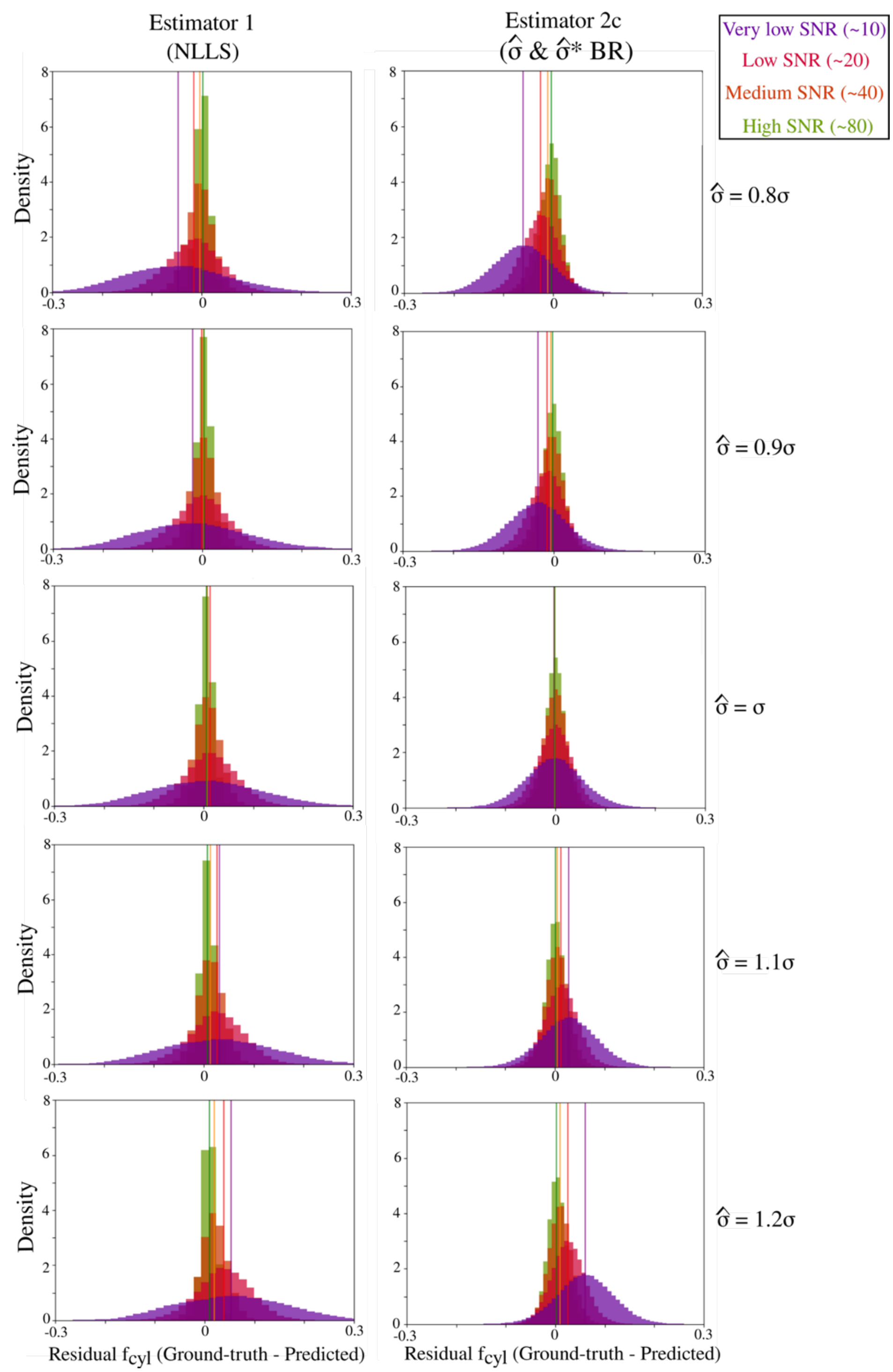

***Figure 7.*** Distribution of $f_{\mathrm{cyl}}$ residuals (ground truth minus estimated) within the white matter across SNR levels (~10, 20, 40, 80) for different levels of noise misestimation during inference. *Estimator 2c* was trained using simulations generated with the scaled noise standard deviation $\hat{\sigma} = \alpha\sigma$, with $\alpha = 0.8,0.9,1.0,1.1,1.2$, while inference was performed on the simulated signal with $\sigma$. Rows correspond to increasing $\alpha$. The vertical colored lines depict the mean of the distribution for the respective SNR level. Correct specification of the noise level ($\hat{\sigma} = \sigma$) yields minimal bias, whereas under- and overestimation introduce systematic deviations that increase in magnitude with both misestimation and decreasing SNR.

*3.4 Performance on in vivo data across SNR levels*

We evaluated the performance of the different estimators on in vivo data acquired at 0.9 $mm^3$ resolution (Figure 8), using both the full dataset averaged across four repeats and a single-repeat acquisition. The four-repeat average yielded an approximate mean SNR of 23.5 (estimated from the non-diffusion-weighted images), whereas the single acquisition had a lower mean SNR of approximately 8.8. The theoretical SNR gain from averaging four repetitions is $\sqrt{4} = 2$, consistent with the observed difference. SNR was estimated using the MPPCA-derived noise standard deviation $\hat{\sigma}^*$ and averaged across the whole brain; therefore, these values represent coarse approximations of the spatially varying true SNR.

At higher SNR (four-repeat average), the parameter maps obtained with the different estimators exhibit only moderate differences (Figure 8, left panel, right column). In contrast, these differences are more pronounced in the lower-SNR single-acquisition data (left panel, left column). The regressor trained on signals with zero-mean Gaussian noise (*Estimator 2b*) exhibits reduced white matter–grey matter contrast and an apparent overestimation of $f_{cyl}$ in white matter. This behavior is consistent with the SNR-dependent Rician-offset bias observed in the simulated experiments (Figures 3–5). In comparison, when accounting (*Estimator 1*, *2a* and *2c*), the white matter–grey matter contrast is preserved, and no systematic overestimation of $f_{cyl}$ in white matter is observed. However, parameter maps estimated by *Estimator 1* visually appear noisier. The proposed RNS approach (*Estimator 2c*) produces more stable parameter maps at low SNR. Moreover, $f_{cyl}$ values remain similar between the high- and low-SNR acquisitions with the lowest mean difference, as shown in the histograms in the right panel of Figure 8, indicating improved robustness to noise-induced bias.

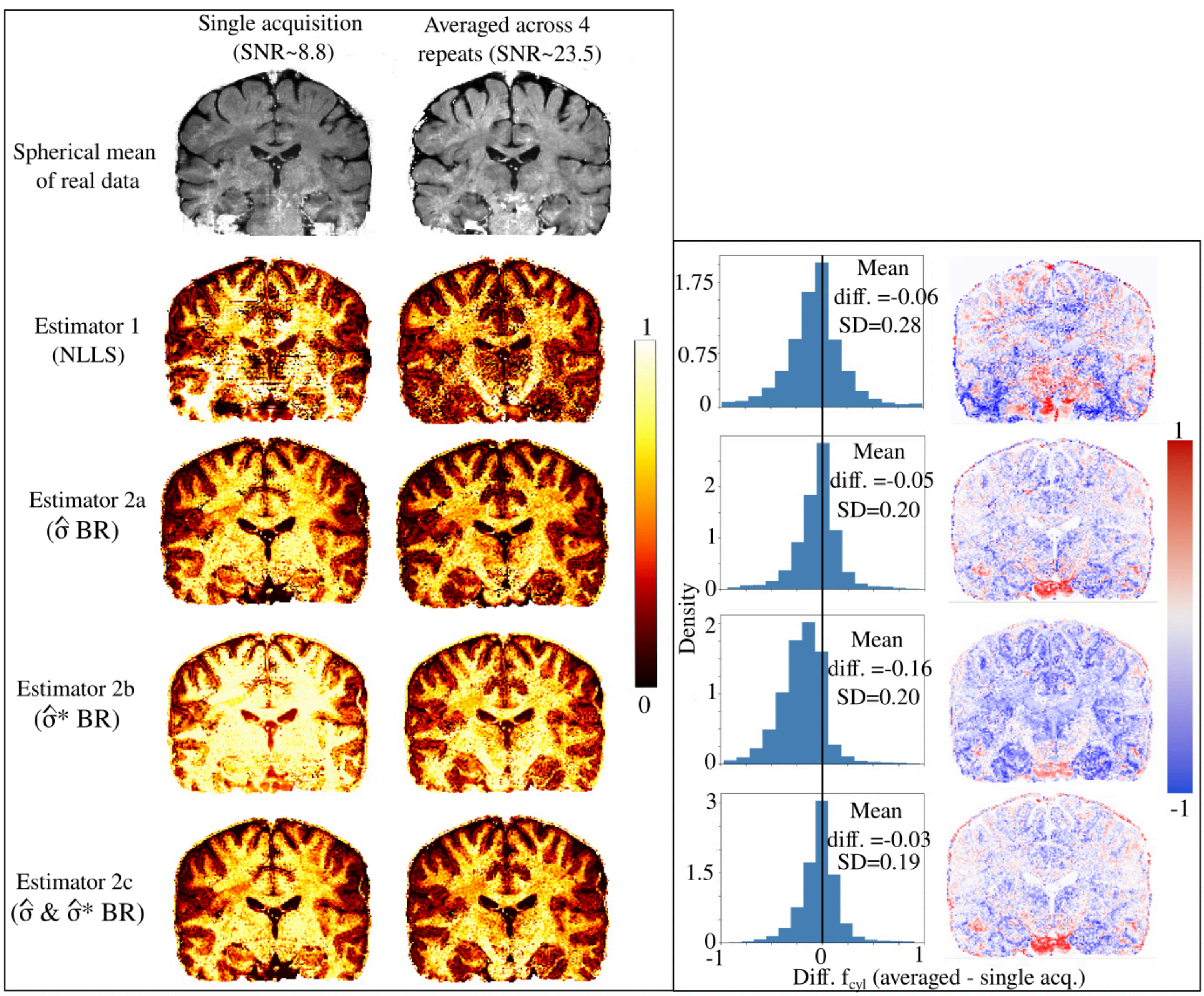

**Figure 8.** Parameter maps of $f_{cyl}$ obtained with the different estimators on the in vivo 0.9 mm³ dataset [71]. In the left panel, the left column shows results from a single acquisition (lower SNR), and the right column shows results from data averaged across four repeats (higher SNR). The top row displays the spherical mean signal at $b = 2$ms/µm² for both the single and averaged acquisitions. The right panel presents histograms of the voxel-wise differences in the estimated parameters between the averaged and single-acquisition data for each estimator shown in the left panel. The solid vertical line indicates zero difference.

*3.5 Dependence of parameter estimation on the regression framework*

To assess whether estimation performance depends on the choice of regression framework, we compared the BR (*Estimators 2a*, *2b* & *2c*) with a MLP (*Estimators 3a*, *3b* and *3c*). Both frameworks were trained using either signals with Rician offset (*2a* & *3a*), zero-mean Gaussian noise (*2b* & *3b*) or signals generated with the proposed RNS framework (*2c* & *3c*).

Figure 9 shows the estimated $f_{\mathrm{cyl}}$ maps at very low (~10) and high (~80) SNRs. Qualitatively, the BR and MLP produce highly similar parameter maps across SNR regimes. Differences between noiseless and noise-informed training dominate over differences between regression architectures.

Figure 10 quantifies the relationship between BR and MLP estimates within white matter across SNR levels. A strong linear correspondence is observed between the residuals of the two frameworks, with comparable bias and standard deviation at each SNR. This indicates that the overall estimation performance is primarily driven by the noise modelling strategy rather than by the specific regression architecture.

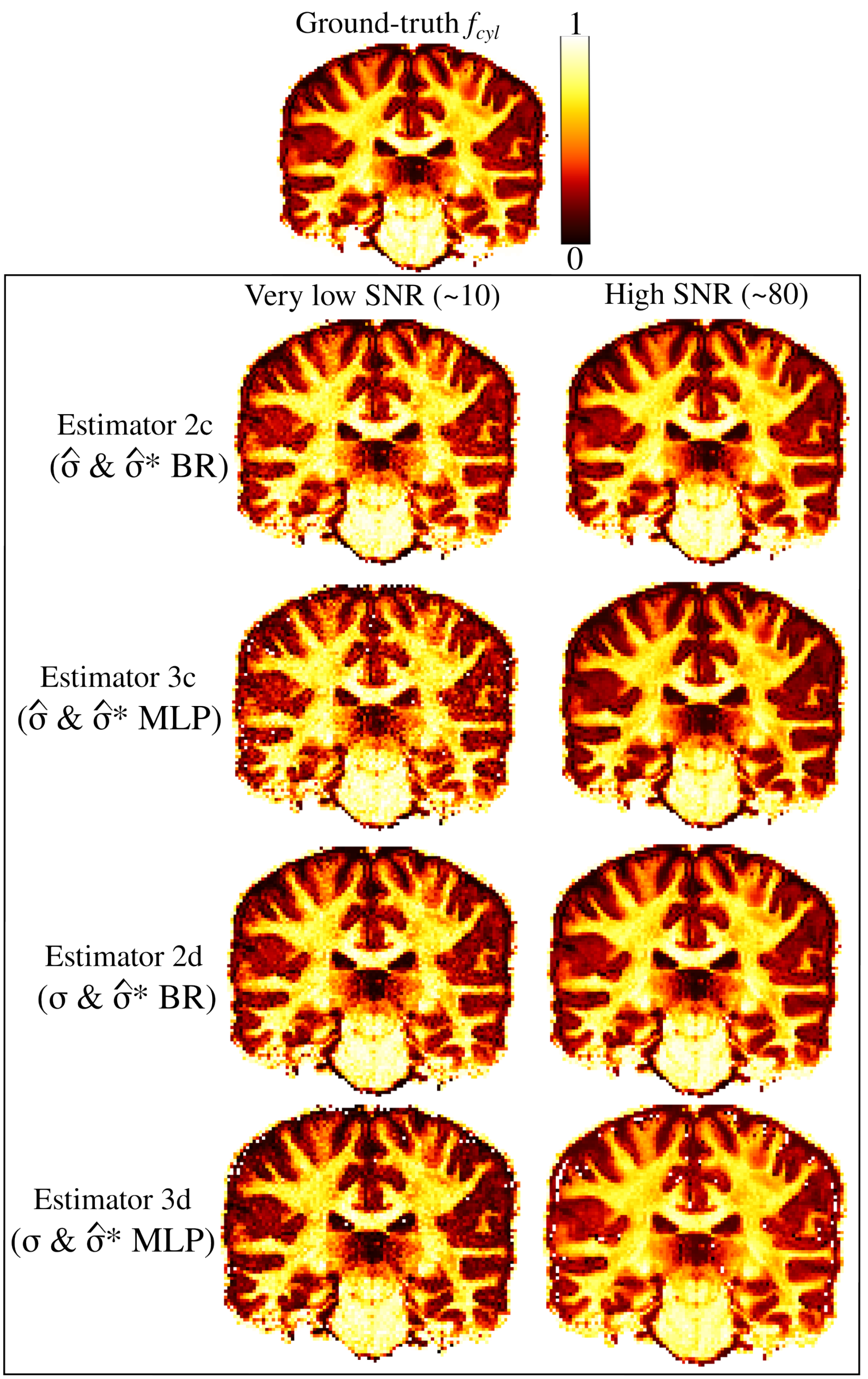

**Figure 9.** Estimated maps from the simulated inference set at very low and high SNR using the BR (*Estimator 2c*) and MLP (*Estimators 3a*, *3b* and *3c*). The ground-truth map is shown at the top for reference. Across SNR levels, maps obtained with the BR and MLP are visually comparable, with differences primarily reflecting the noise modeling strategy rather than the regression architecture.

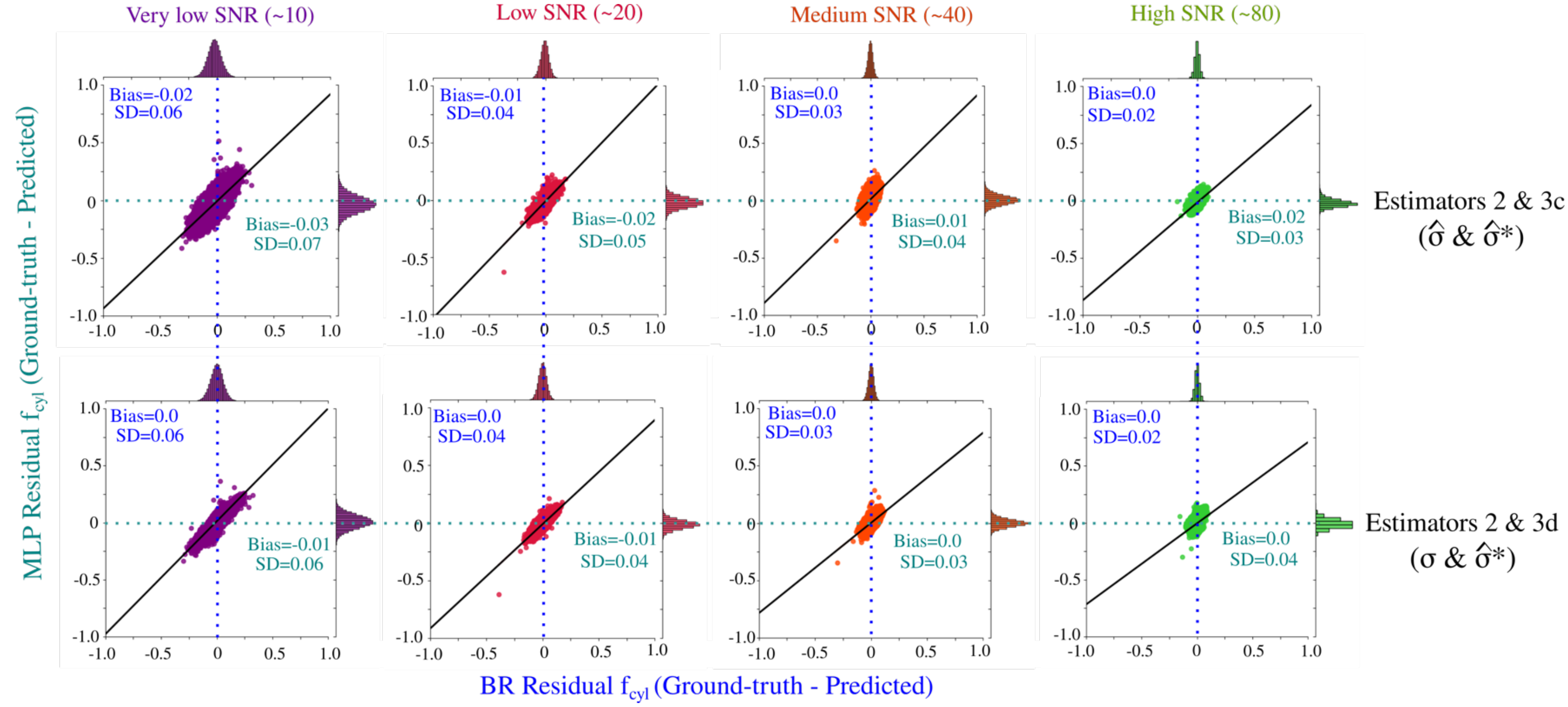

**Figure 10.** Relationship between BR and MLP residuals (ground truth minus estimate) within white matter across SNR levels. Each panel shows the joint distribution of residuals from the BR and MLP; marginal histograms depict the residual distributions for each framework, with mean (bias) and standard deviation (SD) values reported. The dotted blue and teal lines indicate zero residual for the BR and MLP, respectively. The solid black line represents the linear fit between BR and MLP residuals. Across SNR levels, both frameworks exhibit similar bias and dispersion, with a strong linear correspondence between their estimates.

*3.6 Application to the SANDI model*

We next evaluate the proposed framework using the SANDI model, which requires high b-values (up to 6 ms/µm²). The corresponding high b-value signals have intrinsically lower SNR, placing the estimation problem in a regime where noise-induced bias is expected to be more pronounced.

Figure 11 shows parameter maps from a single subject, including the intra-neurite signal fraction ($f_{neurite}$), soma signal fraction ($f_{soma}$), and the soma radius $r_s$. Results are shown for a BR trained on simulated signals with Rician offset (*Estimator 2a*), zero-mean Gaussian noise (*Estimator 2b*) and for the same regressor trained using the proposed RNS framework with $\hat{\sigma}$ and $\hat{\sigma}^*$ (*Estimator 2c*). Across all three parameters, incorporating realistic noise during training leads to improved map quality. For $f_{neurite}$ and $f_{\mathrm{soma}}$, white matter–grey matter contrast and subcortical structures are more clearly delineated with *Estimator 2c*. For $r_s$, *Estimator 2a* and *2b* exhibit widespread extreme values, whereas *Estimator 2c* produces spatially coherent maps with more physiologically plausible parameter ranges. These results are consistent with the improved robustness observed at low SNR in the simulated experiments.

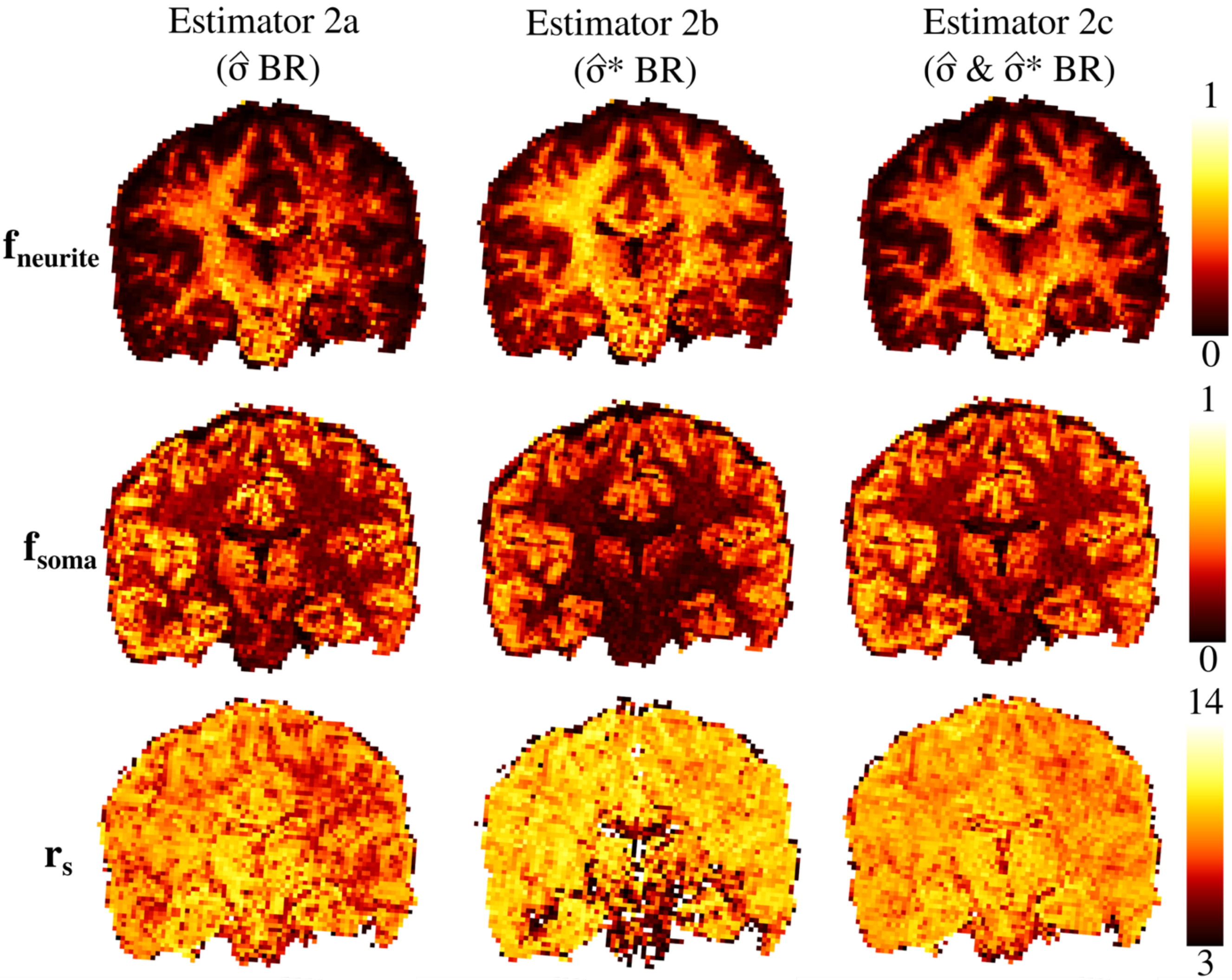

**Figure 11.** SANDI-derived parameter maps from a single subject: intra-neurite signal fraction ($f_{neurite}$), soma signal fraction ($f_{soma}$), and soma radius $r_s$. Estimates are shown for a BR trained on simulated signals with Rician offset (*Estimator 2a*), zero-mean Gaussian noise (*Estimator 2b*) and for the same regressor trained using the proposed RNS framework with $\hat{\sigma}$ and $\hat{\sigma}^*$ (*Estimator 2c*). Incorporating realistic noise during training improves anatomical contrast and reduces extreme parameter estimates, particularly for the soma radius.

## 4. Discussion

In this work, we demonstrate that noise characteristics critically influence microstructural parameter estimation in diffusion MRI using supervised ML, particularly in low-SNR regimes arising for example from high b-values or high spatial resolution. The bias introduced by Rician noise has been described previously in microstructure modeling studies[32,48–50]. Our results extend these observations by showing that, in supervised ML frameworks, explicitly matching the statistical properties of the training data to those of the inference data substantially improves accuracy, precision and robustness to low SNR. Importantly, we show that this effect is largely independent of the regression architecture and is instead driven by how noise is incorporated into the training distribution. We further show that, in real-world applications, supervised ML models accurately trained to reflect the experimental conditions can match the bias performance of classical NLLS, thereby prompting a re-evaluation of previous studies that attributed the higher bias of supervised ML relative to NLLS solely to the sampling strategy used to construct the training set[64,85].

### *4.1 Matching the training and inference distributions*

In supervised ML, performance on unseen data depends critically on how closely the training distribution matches the inference (test) distribution[86,87]. In the present work, *Estimator 2b* was trained on simulated signals corrupted with zero-mean Gaussian noise with standard deviation $\hat{\sigma}^*$, estimated from the corresponding inference dataset. However, this training distribution does not account for the signal-dependent Rician offset inherent to magnitude-reconstructed diffusion data[29,30,40]. As a result, although the variance structure is partially represented, the mean of the training distribution does not match that of the inference data[32,48–50], leading to a residual distributional mismatch. A similar discrepancy arises in *Estimator 2a*, where the Rician expectation is modeled but the effective post-processing standard deviation is not incorporated. Notably, this setting reflects the prevailing practice in supervised learning approaches for dMRI parameter estimation, where simulated training signals typically account for magnitude-induced bias with scalar valued or uniform SNR distributions, but assume simplified or homogeneous noise variance. Our results demonstrate that neglecting the effective post-processing standard deviation introduces a systematic distributional mismatch that becomes a dominant source of bias in low-SNR regimes. This setting corresponds to covariate shift, where the marginal distribution $p(S)$ differs between training and inference domains while the conditional mapping $p(\mathbf{\Theta}|\mathrm{S})$ remains unchanged[86,87]. By explicitly modeling both the magnitude-induced bias and the effective post-processing standard deviation, the proposed RNS framework addresses this fundamental limitation and represents a necessary shift toward statistically consistent training strategies in supervised dMRI parameter estimation.

Such signal-domain mismatch manifests as biased parameter estimates in practice. Indeed, prior studies have reported systematic deviations in supervised diffusion MRI estimators when trained directly on simulation ground-truth values, resulting in biased parameter maps in experimental data[66]. If magnitude-induced bias is not properly modelled during training, supervised estimators may learn a mapping that is inconsistent with the effective signal statistics encountered at inference. Such distributional mismatch particularly increases generalization error and leads to systematic bias in low-SNR regimes, where the Rician expectation departs substantially from the underlying noiseless signal[32,48–50]. The simulated experiments (Figures 3–5) show that *Estimators 2a* & *2b* exhibit marked SNR-dependent bias under these conditions.

Correcting for signal-domain covariate shift is essential for unbiased supervised parameter estimation in diffusion MRI. In this work, we demonstrate that explicitly aligning the statistical properties of the training and inference signals substantially reduces, and in some cases effectively removes, the bias observed in supervised estimators. The proposed RNS framework can be interpreted as one practical implementation of such a covariate shift correction in the signal domain, aligning the training distribution of the simulated signals with the inference distribution of the acquired signals, to reduce generalization error[86]. Specifically, the RNS approach introduces a two-step strategy (*Section 2.2*): first, applying the expectation of a Rician distribution using an estimate $\hat{\sigma}$ derived from the inference dataset to model the signal-dependent offset; and second, adding zero-mean Gaussian noise with standard deviation $\hat{\sigma}^*$, estimated from spherical harmonic residuals, to reproduce the effective post-processing standard deviation. The resulting synthetic training data are therefore Gaussian distributed with a mean governed by the Rician expectation and a standard deviation consistent with the inference set (Figure 2).

The impact of this alignment is reflected across both simulated and in vivo experiments. In the simulated inference set, incorporating the Rician offset (*Estimators 1, 2a, 2c* & *2c*) largely removes systematic bias across SNR levels (Figure 5), while additionally modelling the standard deviation structure (*Estimators 2b, 2c* & *2d*) reduces residual dispersion. In the real 0.9 mm³ dataset, which has lower intrinsic SNR, the RNS-trained estimator produces more stable white matter–grey matter contrast and reduces the apparent overestimation of $f_{\mathrm{cyl}}$ compared to training on signals corrupted with zero-mean Gaussian noise (Figure 8). Similar improvements are observed for the higher-order SANDI model at high b-values (Figure 11), where low SNR amplifies noise-induced bias.

It is worth noting that the improvements observed here are not specific to the BR architecture. As shown in *Section 3.5* (Figures 9–10), comparison with a MLP yielded highly similar behavior across SNR regimes, including comparable bias and standard deviation characteristics. This indicates that the dominant factor governing estimation performance in this setting is the noise modelling strategy rather than the specific regression architecture. Although more expressive learning frameworks may provide incremental improvements, architectural complexity alone cannot compensate for a mismatch between the statistical properties of the training and inference signals. These results emphasize that accurate modeling of the input signal distribution is a prerequisite for mitigating systematic noise-induced bias.

*4.2 Noise modelling versus parameter distribution design in supervised training*

Prior work has emphasized that the performance of supervised learning approaches for diffusion MRI depends strongly on the design of the training set, particularly the distribution of model parameters used to generate simulated signals. For example, sampling parameters from uniform versus Gaussian distributions has been shown to affect estimation accuracy and precision[64]. Similarly, recent work has highlighted the importance of label design and training targets, demonstrating that the choice of parameter labels themselves can influence estimator behaviour and bias[66]. These studies primarily focus on the distribution of the latent variables $\mathbf{\Theta}$, and thus on alignment of the parameter space between training and inference domains.

In contrast, the present work addresses a distinct but complementary source of mismatch: the statistical distribution of the input signals $S$ [87]. Even when parameter sampling is biologically plausible and label design is carefully considered, a discrepancy between the noise characteristics of simulated training signals and those of experimentally acquired data induces a covariate shift in signal space. In this setting, the conditional mapping $p(\mathbf{\Theta}|S)$ is assumed invariant, but the marginal distribution $p(S)$ differs between training and inference due to magnitude-induced bias and post-processing standard deviation. Our results demonstrate that correcting this signal-domain mismatch substantially reduces systematic bias, independent of how parameters are sampled.

These observations highlight that parameter distribution design and signal noise modelling operate along distinct axes of the learning problem. Parameter sampling governs coverage of the model manifold, whereas noise modelling determines statistical alignment between simulated and acquired signals. The present findings suggest that realistic modelling of signal noise characteristics is a critical prerequisite for unbiased supervised estimation and complements careful parameter and label design.

*4.3 Noise estimation and practical considerations*

A key practical factor is the accuracy of the noise estimate. The controlled misestimation experiments (Figure 7) demonstrate that even moderate under- or overestimation of $\sigma$ (±10–20%) introduces systematic bias in the recovered parameters, with effects that become increasingly pronounced at low SNR. When $\sigma$ is underestimated, the Rician expectation correction is insufficient, resulting in negative residual bias (i.e. an underestimation of $f_{cyl}$). Conversely, overestimation of $\sigma$ induces a positive bias (i.e. an overestimation of $f_{cyl}$). These findings are consistent with the differences observed between using ground-truth $\sigma$ and estimated $\hat{\sigma}$ (Figures 4 & 5), and with our observation that MPPCA-based estimates tend to underestimate the true noise level (Supplementary Figure S1), as reported previously. Together, these results indicate that misestimation of $\sigma$ directly propagates into parameter bias through the signal-dependent Rician correction. This underscores the importance of robust and accurate noise estimation, particularly for supervised ML frameworks that explicitly model noise statistics during training and inference. In practice, slightly lower estimates are considered safer[88].

In many supervised learning approaches for dMRI, noise is introduced by sampling the noise standard deviation uniformly within a predefined range that encompasses plausible experimental values. While this strategy aims to improve robustness to varying noise conditions, it does not ensure that the training data reflect the actual noise distribution of the target dataset. In practice, this mismatch can lead to underrepresentation of the most relevant noise regimes, effectively reducing the useful sample size and requiring larger training datasets to achieve adequate coverage of the inference distribution[89]. By contrast, the proposed RNS framework samples noise parameters directly from empirically estimated distributions, ensuring that the most representative noise levels are appropriately captured during training. This results in a closer alignment between training and inference signals without the need for substantially increasing the size of the training dataset.

We estimated $\hat{\sigma}^*$ from SH residuals aggregated across shells (Supplementary Figure S1). Other strategies, such as wavelet-based approaches or shell-specific estimators, may provide improved estimates. The choice of SH truncation level ($l_{max}$) also influences the residual structure and therefore the inferred noise variance. Future work should systematically evaluate alternative noise estimation schemes and their impact on parameter inference.

From a preprocessing perspective, methods such as MPPCA render magnitude data approximately Gaussian. However, residual signal-dependent bias may persist, especially at low SNR or high b-values. Access to complex data for denoising would theoretically allow removal of Rician bias more completely[90], potentially making noiseless training data a better match to the inference signal. In practice, complex data are rarely available, and magnitude-based preprocessing remains the norm.

An alternative strategy would be to denoise the acquired data such that the noiseless simulated signal becomes an adequate approximation of the inference set[88]. Conceptually, this is equivalent to moving the inference distribution toward the training distribution rather than vice versa. Both strategies aim to reduce domain mismatch; which approach is preferable may depend on data availability and preprocessing constraints.

*4.4 Limitations*

A primary limitation of the proposed framework lies in its reliance on accurate estimates of the noise statistics. The RNS approach requires reliable estimation of both the thermal noise standard deviation $\hat{\sigma}$ and the effective post-processing noise standard deviation $\hat{\sigma}^*$. In this work, $\hat{\sigma}$ was estimated using MPPCA, which tends to underestimate the true noise level (Supplementary Figure S1), and as shown in our experiments, such misestimation directly propagates into parameter bias through the Rician expectation. The estimation of $\hat{\sigma}^*$ from spherical harmonic (SH) residuals also depends on factors such as the angular sampling within each b-value shell and the choice of truncation order $l_{\max}$. While our results (Supplementary Figure S1) support the validity of this approach, alternative estimators may provide improved robustness.

Another limitation is that the proposed RNS framework was primarily evaluated on spherical-mean signals, where directional dependencies are averaged out and the noise characteristics are simplified. Extending this approach to fully directional signals presents additional challenges, particularly in highly anisotropic tissues such as white matter. In such settings, the interaction between noise, orientation dispersion, and model nonlinearity can lead to more complex bias patterns that are not fully captured by the present framework [91]. Addressing these challenges will require further investigation into noise modelling strategies that account for directional signal structure.

## Conclusion

We demonstrate that explicitly incorporating realistic noise statistics into simulated training data substantially improves the robustness and accuracy of supervised microstructure parameter estimation, particularly in low-SNR regimes. These improvements arise from reducing distributional mismatch between training and inference signals, thereby lowering epistemic uncertainty and mitigating systematic bias. Our results indicate that careful modeling of magnitude-induced bias and effective post-processing noise standard deviation is a critical component in the design of supervised diffusion MRI inference frameworks. This consideration becomes especially important for advanced microstructure models and acquisition protocols that inherently operate in low-SNR conditions (e.g. high-resolution data).

## Acknowledgements

B.G.K. is supported by a Natural Science and Engineering Research Council of Canada Postdoctoral Research Award (NSERC-CPRA).

J.V. is supported by NIH grants R21AG087904, P41EB017183 and the National Institute on Aging of NIH (R21AG083539).

M.J. and M.P. are supported by UKRI Future Leaders Fellowship (MR/T020296/2 and 1073).

A.R.K. is supported by the Canada Research Chairs program #950-231964, NSERC Discovery Grants RGPIN-2015-06639 and RGPIN-2023-05558, Canadian Institutes for Health Research Project grant #366062, Canada Foundation for Innovation (CFI) John R. Evans Leaders Fund project #37427, the Canada First Research Excellence Fund, and Brain Canada.

S.A.-F. is supported by the research grant PID2021-124407NB-I00, funded by MCIN/AEI/10.13039/501100011033 and the European Union “NextGenerationEU/PRTR”, the research grant VA156P24 funded by Junta de Castilla y León (Consejería de Educación) and the European Regional Development Fund (ERDF/FEDER), and the research grant PRX24/00161 (Estancias de movilidad, modalidad senior) funded by Ministerio de de ciencia, innovación y universidades.

**Supplementary Figures**

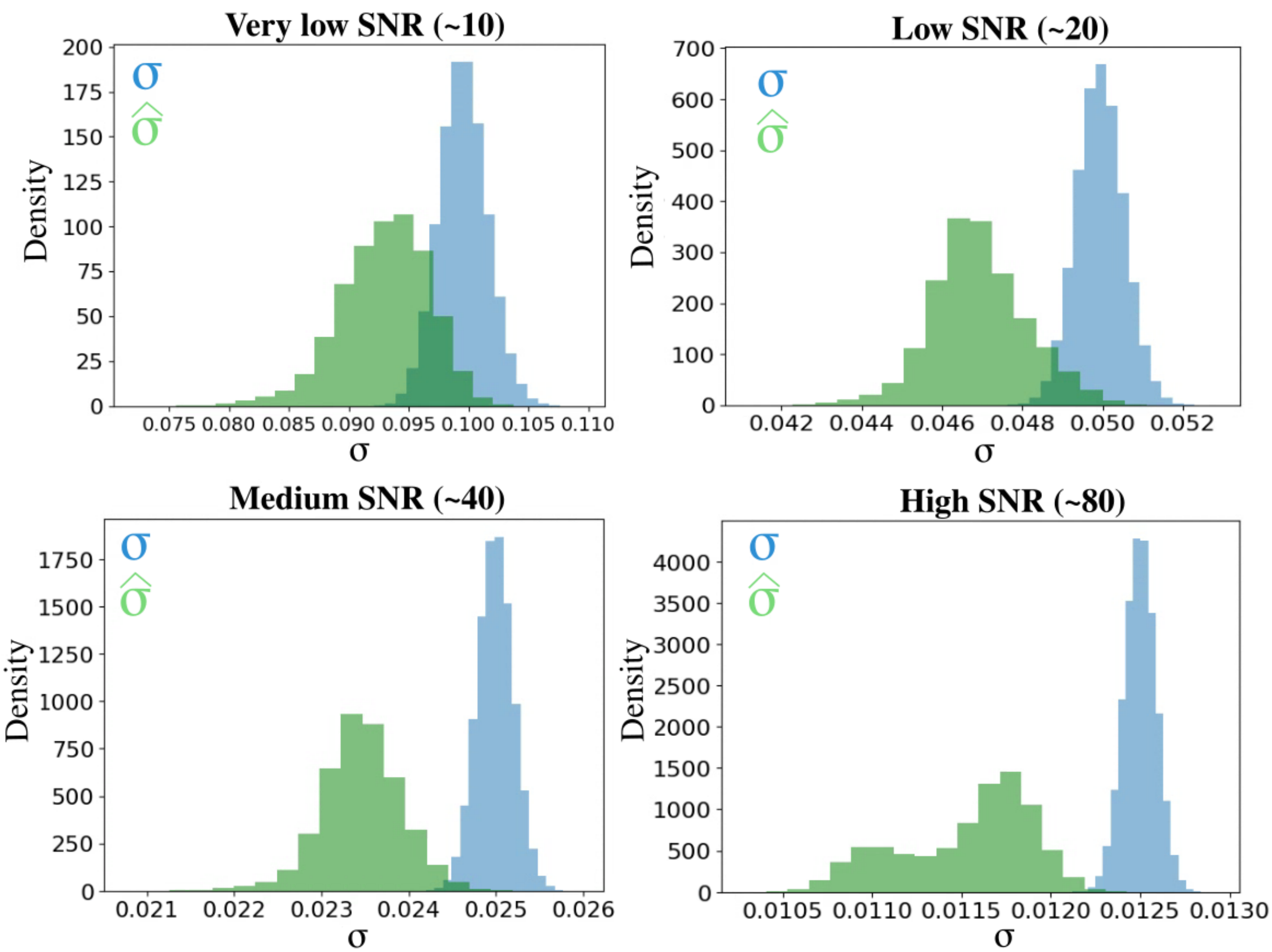

**Supplementary Figure S1.** Comparison of the ground-truth distribution of $\sigma$ (blue) in the brain and $\hat{\sigma}$ estimated from MPPCA (green) across SNR levels.

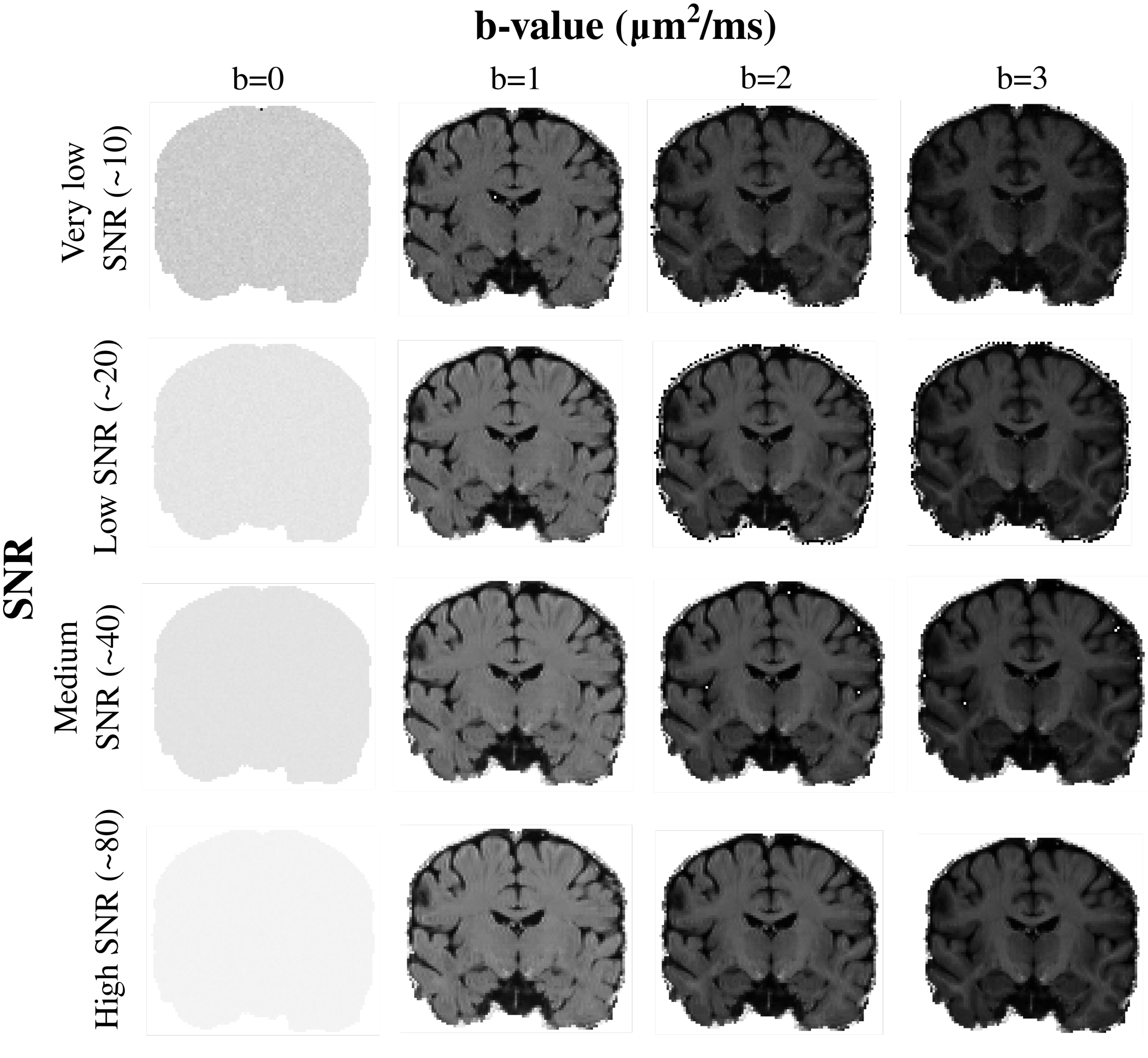

**Supplementary Figure S2.** Simulated normalized spherical mean signals across SNR and b-value (see *section 2.4*).

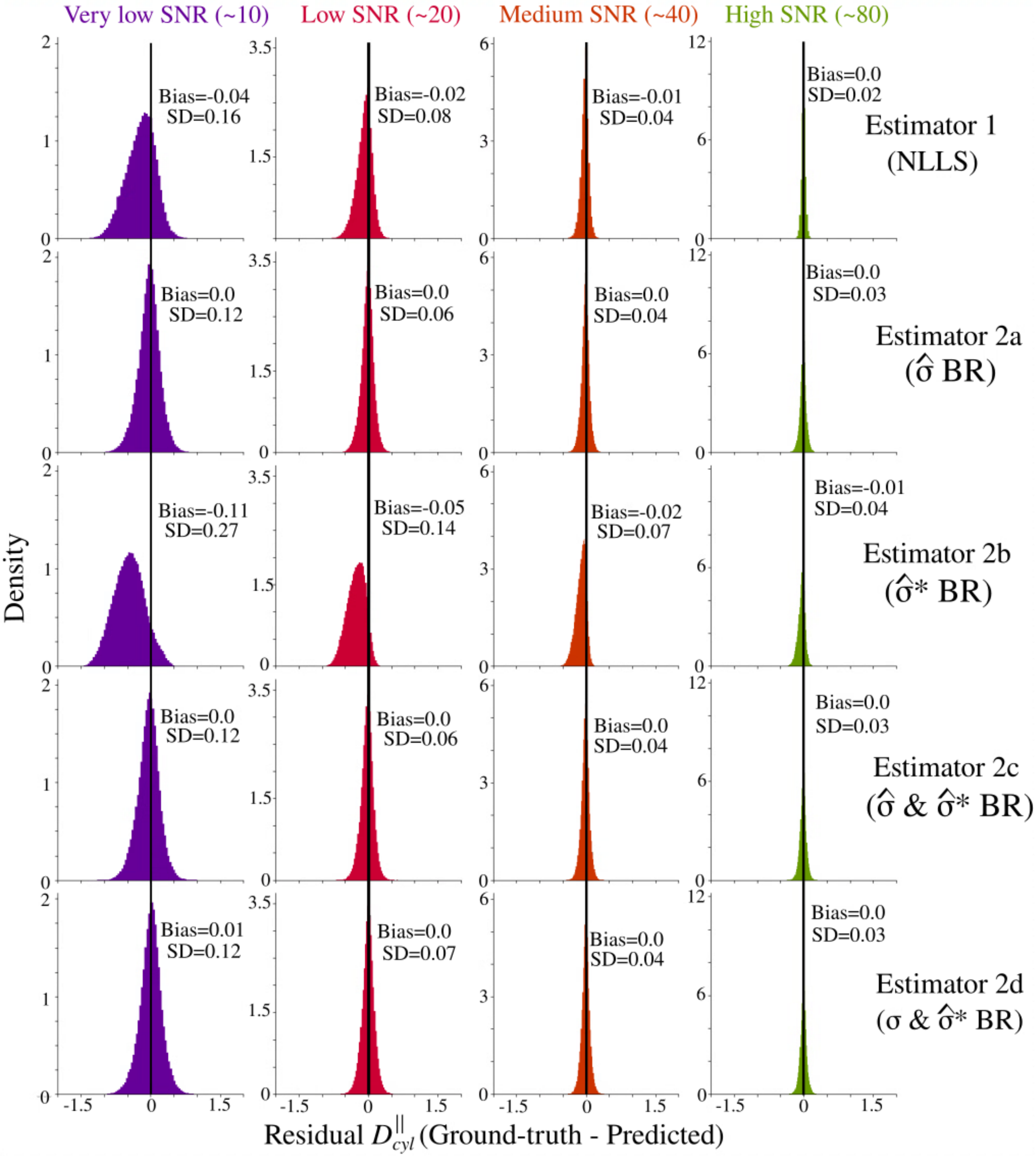

**Supplementary Figure S3.** Distribution of $D_{cyl}^{||}$ residuals (ground truth minus estimate) within the white matter across SNR levels for each estimator. Columns correspond to increasing SNR (~10, 20, 40, 80), and rows to estimators. The vertical black line indicates zero residual. At high SNR, all estimators exhibit low bias and variance. At lower SNR, bias emerges when Rician effects are not modeled (*Estimator 2b*), whereas estimators incorporating Rician noise modeling (*Estimators 1*, *2a*, *2c*, *2d*) maintain near-zero bias.